\documentclass[a4paper,10pt]{article}

\usepackage{PRIMEarxiv}
\usepackage{mathpazo}
\usepackage[utf8]{inputenc} 
\usepackage{hyperref}
\usepackage{todonotes}
\usepackage{amsfonts} 
\usepackage{amsmath}
\usepackage{multirow}
\usepackage{array}
\usepackage{listings}
\usepackage{eurosym}
\usepackage{cleveref}
\usepackage{ragged2e}
\usepackage{makecell}
\usepackage{booktabs}
\usepackage[labelfont=bf, size = small, textfont=it]{caption}
\usepackage{floatrow}
\Crefformat{equation}{Equation~#2#1#3}

\author{
    Marco Bronzini \\
    University of Trento, Ipazia S.p.A. \\
    Trento, Milano (Italy) \\
    \texttt{marco.bronzini-1@unitn.it} \\
   \And
    Carlo Nicolini \\
    Ipazia S.p.A. \\
    Milano (Italy) \\
    \texttt{c.nicolini@ipazia.com} \\
   \And
    Bruno Lepri \\
    Fondazione Bruno Kessler (FBK), Ipazia S.p.A.\\
    Trento, Milano (Italy) \\
    \texttt{lepri@fbk.eu} \\
   \And \hspace{-12mm}
    Andrea Passerini \\ \hspace{-12mm}
    University of Trento\\ \hspace{-12mm}
    Trento (Italy) \\ \hspace{-9mm}
    \texttt{andrea.passerini@unitn.it} \\
   \And
    Jacopo Staiano \\
    University of Trento\\
    Trento (Italy) \\
    \texttt{jacopo.staiano@unitn.it} \\
}

\title{Glitter or Gold? Deriving Structured Insights from Sustainability Reports via Large Language Models}

\begin{document}
\maketitle

\begin{abstract}
Over the last decade, several regulatory bodies have started requiring the disclosure of non-financial information from publicly listed companies, in light of the investors' increasing attention to Environmental, Social, and Governance (ESG) issues.
Publicly released information on sustainability practices is often disclosed in diverse, unstructured, and multi-modal documentation. This poses a challenge in efficiently gathering and aligning the data into a unified framework to derive insights related to Corporate Social Responsibility (CSR).
Thus, using Information Extraction (IE) methods becomes an intuitive choice for delivering insightful and actionable data to stakeholders.
In this study, we employ Large Language Models (LLMs), In-Context Learning, and the Retrieval-Augmented Generation (RAG) paradigm to extract structured insights related to ESG aspects from companies' sustainability reports.
We then leverage graph-based representations to conduct statistical analyses concerning the extracted insights.
These analyses revealed that ESG criteria cover a wide range of topics, exceeding 500, often beyond those considered in existing categorizations, and are addressed by companies through a variety of initiatives.
Moreover, disclosure similarities emerged among companies from the same region or sector, validating ongoing hypotheses in the ESG literature.
Lastly, by incorporating additional company attributes into our analyses, we investigated which factors impact the most on companies' ESG ratings, showing that ESG disclosure affects the obtained ratings more than other financial or company data.
\end{abstract}



\section{Introduction}  \label{sec:intro}
Public health, climate change, social inequalities, diversity, and inclusiveness are challenges that need global attention as well as innovative and collaborative solutions.
However, building a sustainable society requires defining a common set of sustainable-related issues to disclose, measure and comply with.
ESG, which stands for Environmental, Social, and Governance, is an established set of principles used to monitor the sustainability and ethical practices of businesses within society.
These three E/S/G aspects are further described via more granular indicators (both qualitative and quantitative) concerning, for example, waste management, emissions, labour rights, and diversity. 
These indicators aid in evaluating the degree to which a corporation contributes to achieving societal goals.
Assessing these ESG aspects can also help monitor the progress of the seventeen Sustainable Development Goals (SDGs) included in the United Nations' 2030 Agenda for Sustainable Development~\cite{unSustainableDevelopment} which sets ambitious goals for building a sustainable society such as gender equality, responsible consumption and production, and climate action.
Over the last decade, there has been a growing demand for disclosing companies' non-financial information. 
This demand comes from legislation such as the European Union's Non-Financial Reporting Directive (NFRD,~\cite{nonFinancialReportingDirective}) which requires all public-interest companies with more than 500 employees to to do so.
The more recent European Union's Corporate Sustainability Reporting Directive (CSRD,~\cite{corporateSustainabilityReportingDirective}) further increases this demand by enlarging the pool of companies concerned by a factor of 4: from roughly 12 thousand to 50 thousand companies~\cite{corporateSustainabilityReportingDirective}.

Non-financial disclosures are typically reported in sustainability reports, web pages, social media posts, news, press releases or earning calls. 
To overcome this variety of sources, stakeholders generally rely on a third-party assessment of corporations’ ESG performance to inform their decisions: ESG ratings provided by agencies such as \emph{Sustainalytics}, \emph{MSCI}, \emph{S\&P Global}, \emph{Moody’s} and \emph{Refinitiv}~\cite{RateRaters2020}. 
These rating agencies rely on proprietary assessment methodologies with different perspectives on the measurement, scope and weight of different ESG aspects. 
This creates divergences in companies' evaluations across agencies and thus unsatisfactory degrees of explainability, transparency or fairness~\cite{chatterji2016ratings, abhayawansa2021sustainable, billio2021inside, berg2022aggregate}. 
Stakeholders might overcome this issue by directly accessing non-financial information and imposing their scope and weight to assess corporates' ESG performance~\cite{RateRaters2020, ehlers2023deconstructing}. 
However, extracting meaningful insights from ESG-related data sources can be challenging and laborious, often including lengthy documents.
Consequently, the stakeholders can face significant obstacles in evaluating companies' ESG performance from opaque and divergent assessments to verbose textual documents.
We posit that a data-driven approach, coupled with state-of-the-art Natural Language Processing (NLP) techniques, can provide automatic tools to extract insights from companies' disclosures such as sustainability reports. 
Further, the proposed methodology allows us to better understand the companies' ESG assessments and to unveil relationships between what companies disclose and their ratings.
 
The purpose of this work is to automatically extract and analyse the ESG initiatives disclosed by companies in their sustainability reports, and to investigate how these impact their ESG performance assessment.
Our proposed methodology relies on Large Language Models (LLMs,~\cite{brown2020language, wei2021finetuned, wei2022emergent}) for Information Extraction purposes, on graph-based representations for data analysis, and on the SHapley Additive exPlanations (SHAP,~\cite{NIPS2017_7062}) framework for the interpretability of the ESG scores. 
Specifically, we employed a generative LLM to extract structured insights from companies' sustainability reports, bipartite graphs\footnote{A bipartite graph is a graph whose nodes can be divided into two distinct and independent partitions~\cite{asratian1998bipartite, guillaume2006bipartite}} to conduct analyses on them, and the SHAP framework on linear regression to investigate the impact of companies' disclosures on ESG scores.
Turning the unstructured information from sustainability reports into a structured and unified format allows us to build graph-based representations; these, in turn, are directly usable and exploitable for a diverse set of tasks, from exploring and navigating the data to discovering emerging patterns via statistical and interpretability analyses. 

As mentioned before, for this first Information Extraction step, we employ an instruction-finetuned generative LLM (WizardLM~\cite{xu2023wizardlm}), leveraging both In-Context Learning (ICL~\cite{dong2023survey}) and Retrieval Augmented Generation (RAG,~\cite{lewis2020retrieval}), to extract ESG-related actions as triples from companies' sustainability reports.
LLMs have consistently been shown to hold semantic understanding and store factual knowledge~\cite{lewis2020retrieval, kojima2022large, trott2023large, zhao2023survey};
Instruction-tuning (that is, providing task descriptions via natural language instructions) further enhances their capabilities to address downstream NLP tasks, such as Information Extraction~\cite{brown2020language, reynolds2021prompt, wang2023aligning, xu2023wizardlm, zhang2023instruction};
LLMs have also demonstrated a remarkable ability as few-shot learners using In-Context Learning~\cite{dong2023survey}, a technique that relies on providing a few input-output samples within the model instruction. 

The triple format allows us to represent the document sentences in a standardised format following a pre-defined semantic template tailored to ESG aspects.
These extracted triples are then used to build a Knowledge Graph (KG), following the same research direction of recent studies based on OpenAI’s commercial LLMs~\cite{carta2023iterative, meyer2023llm, trajanoska2023enhancing, zhu2023llms}.
This allows us to condense all the companies' disclosures through a graph representation which offers a versatile approach to illustrate structured information~\cite{reinanda2020knowledge} as concepts (nodes) and their relationships (edges)~\cite{fensel2020introduction, hogan2021knowledge}.
The generated KG is then decomposed into several bipartite graphs, thus two-fold graph representations, to analyse the companies' disclosures by inspecting the extracted information from various angles, including the company and topic perspective.

Our findings include descriptive, similarity and correlation analyses concerning ESG-related actions disclosed by companies in their sustainability reports.  
These analyses unveiled that companies addressed ESG topics from several perspectives, spotlighting the complexity of ESG-related matters.
In addition, similarities in companies' disclosures emerged from companies from the same geographical region and the same sector, remarking the possible influence of exogenous factors~\cite{yu2021international, baldini2018role} and the presence of sector-focused topics~\cite{eccles2012need, khan2016corporate, busco2020preliminary}.
Further, our interpretability analysis of ESG scores highlights how a company's disclosures impact its ESG rating more than other company-specific or financial aspects.
Finally, our analyses show the rewards of transparency: comprehensive disclosure of non-financial information appears to affect ESG scores positively, whereas, conversely, reporting on a limited set of ESG categories seems detrimental;
interestingly, we also observe a significant incidence of other factors not directly related to ESG, such as a company's incorporation year and continent (Europe).

Overall, our work contributes to the literature on sustainability and Corporate Social Responsibility (CSR) by proposing an advanced NLP pipeline for automatically extracting information from companies' sustainability reports 
and validating some ongoing hypotheses using a data-driven approach and the company perspective.


\section{Related work} \label{sec:relatedWork}
In this section, we first discuss the state-of-the-art approaches for creating Knowledge Graphs (KGs, \Cref{subsec:rwKgGeneration}), encompassing a spectrum ranging from conventional NLP pipelines to the exploitation of Large Language Models (LLMs).
In \Cref{subsec_lwTextAnalysics}, we then summarise the main studies focused on ESG-related textual information.

\subsection{Knowledge Graphs generation} \label{subsec:rwKgGeneration}
Knowledge Graphs (KGs,~\cite{fensel2020introduction, hogan2021knowledge}) offer a versatile method of representing knowledge that can be leveraged in various use cases and domains~\cite{zou2020survey, ji2021survey}. 
They can be applied to question-answering~\cite{jia2021complex}, recommendation systems~\cite{shao2021survey}, and information retrieval~\cite{reinanda2020knowledge}. 
The task of KG generation, also known as knowledge acquisition, aims to create KGs by extracting information from unstructured, semi-structured or structured sources as well as augmenting existing graphs~\cite{ji2021survey, fensel2020knowledge,yan2018retrospective}.
Traditional approaches for knowledge acquisition involve several NLP tasks which are generally disjointly learned, a process that is prone to error accumulation.
To overcome this problem, new one-stage NLP pipelines have been proposed to jointly extract both entities and relations~\cite{katiyar2017going, wang2020tplinker}. 

In this context, Open Information Extraction (OIE,~\cite{niklaus2018survey}) emerged as the task of extracting structured information, typically in the form of subject-predicate-object (SPO) triples, without relying on a predefined template or a specific domain. 
These SPO triples can then be leveraged to generate knowledge graphs based on the subjects, predicates and objects extracted from textual documents.
This approach can mitigate the impact of depending on external knowledge, such as patterns and domain-specific heuristic rules present in the training data. 
Recently, OIE models (e.g., Multi$^2$OIE~\cite{ro2020} and DeepEx~\cite{wang2021}) have employed transformer-based LLMs (e.g., BERT~\cite{devlin2018bert}) to extract both syntactic and semantic
information~\cite{liu2022open}.

LLMs trained on large-scale corpora have demonstrated significant
potential across diverse NLP tasks thanks to the technique of
prompt engineering~\cite{pan2023unifying}.  
Many researchers~\cite{carta2023iterative, meyer2023llm, trajanoska2023enhancing, zhu2023llms} have demonstrated the ability
of LLMs (i.e., OpenAI's GPT models) to extract structured data from
texts, accomplishing the KG generation task. Furthermore, LLMs can
enhance KGs through several other perspectives: (1) enrich entity and
relation representations using embeddings, (2) generate new facts
(i.e., KG completion~\cite{chen2023knowledge}), (3) produce
natural-language descriptions of KG facts (i.e.,
KG-to-text~\cite{axelsson2023using}) and (4) answer natural-language
questions (i.e., question-answering~\cite{pan2023unifying}).

We follow this promising research direction by exploiting the semantic understanding, generative abilities and flexibility of LLMs to extract ESG-related structured information.
Our methodology adopts a Large Language Model, alongside the Retrieval Augmented Generation (RAG) paradigm~\cite{lewis2020retrieval} and the In-Context Learning (ICL) technique, to overcome the main limitation of conventional OIE methods in achieving our goal of extracting ESG-related structured insights:
OIE methods traditionally extract structured information by relying only on the syntactical structure of a sentence, without any pre-defined template, which poses important limitations in the type of information extracted.
They may fail to extract, for example, information related to specific domains or activities that are not the direct subject of the sentence.
An information extraction approach based on generative LLMs allows us, instead, to generate semantically-aware and ESG-oriented triples instead of traditional SPO triples, enabling the creation of a full-fledged ESG-oriented KG.

\subsection{Text Analytics on ESG-related information} \label{subsec_lwTextAnalysics}

Several works have explored the use of NLP technology to 
process companies' non-financial textual information and extract
meaningful insights concerning statements, facts and actions disclosed
by companies. For example, Chou \emph{et al.}~\cite{chou2023firms}
applied distributional semantics on 10-K filings (i.e., annual
financial reports required by the U.S. Securities and Exchange
Commission) for extracting topics related to climate change that are
disclosed by companies, aiming to monitor environmental policy
compliance. Raghupathi \emph{et al.}~\cite{raghupathi2020identifying}
applied traditional approaches ranging from trigram co-occurrences to
clustering and classification to gain insights into the shareholders’
perspectives and objectives regarding sustainability and climate
change. 
LLMs have also been leveraged for Text Analytics in the ESG domain.
Jacouton \emph{et al.}~\cite{jacouton2022proof,sdgprospectorProspectorAgence}
introduced \textit{SDG Prospector}, a tool exploiting LLMs to identify
paragraphs in Public Development Banks' sustainability reports that
address SDGs.
Similarly, Webersinke \emph{et al.}~\cite{webersinke2021climatebert} introduced \textit{ClimateBert}, a transformer-based language model fine-tuned for climate-related classification tasks.
Vaghefi \emph{et al.}~\cite{vaghefi2023chatclimate} adopted the
paradigm of RAG~\cite{lewis2020retrieval} to enrich GPT-4 with the
ability to reliably answer climate-related questions, by augmenting
the posed questions with contextual information retrieved from the
Sixth Assessment Report, released by the United Nations
Intergovernmental Panel on Climate Change (IPCC). The authors also
released a conversational agent~\cite{chatclimateChatClimate} based on
their proposed approach. Ni \emph{et al.}~\cite{ni2023paradigm},
developed \textit{ChatReport}, an LLM-based tool that evaluates
companies' sustainability reports according to the eleven
recommendations~\cite{board2017recommendations} provided by the Task
Force for Climate-Related Financial Disclosures (TCFD). The tool
combines semantic search to identify text chunks that are pertinent to
each recommendation and LLM prompting to summarize them.

Our work differs from existing approaches by jointly (i) leveraging generative LLMs for Knowledge Graph (KG) generation,  
(ii) employing an open-sourced generative LLM and (iii) relying directly on companies' sustainability reports.
This methodology, alongside the exploitation of bipartite graph representation, allows us to conduct non-trivial analyses concerning ESG categories and actions disclosed by companies.

\section{Materials and Methods} \label{sec:materialsMethods}
This section discusses the data, the approaches, and the methods used in this work. First, we describe the data sources (\Cref{subsec_dataSources}); second, we provide a detailed overview of our approach, from data preparation (\Cref{subsec:dataPreparation}) to triple generation (\Cref{subsec:RetrievalAugmentedGeneration}) and KG generation (\Cref{subsec:knowledgeGeneration}).
Finally, we discuss the methods and approaches used to analyse, compare and evaluate the findings concerning the generated triples (\Cref{subsec:analysisMethods}).


\subsection{Data sources} \label{subsec_dataSources}
Here, we describe the three main data sources used in our work, which include (i) sustainability reports (\Cref{subsubsec:sustainabiliyReports}), (ii) an ESG categorization (\Cref{subsubsec:esgCategorization}), and (iii) ESG rating scores and other companies' information (\Cref{subsubsec:companyInformation}).

\subsubsection{Sustainability reports} \label{subsubsec:sustainabiliyReports}
Sustainability reports are non-financial documents published by companies to disclose information concerning the impact of their activities on the environment and people. 
Therein are described the actions the company took or expects to take regarding ESG matters -- such as respect for human rights, fair treatment of employees, anti-corruption and bribery as well as board diversity~\cite{corporateSustainabilityReportingDirective}.
However, ESG reporting can be subjective and opaque due to the complexity of reporting qualitative aspects, particularly those related to social and governance issues~\cite{doyle2018ratings}.
Furthermore, the lack of a standardised framework for ESG reporting makes quantitative/comparative analyses difficult~\cite{oecd_sustainable_2020, aliakbari2023impracticality}.

We initially collected 6,456 sustainability reports from 4,222 different companies using the report URLs available on two public websites~\cite{sasbReports, responsibilityReports}.
Although these sustainability reports are mainly written in English (94\% of all the available documents), the nationality of the companies is fairly diversified, covering 74 different countries. 
However, the majority of the available reports (56\%) come from North American companies.
For our study, we consider only the reports written in English because of its broad coverage and the wide range of pre-trained language models available for this language (many thousands of language models on the Hugging Face platform \cite{HuggingFaceLanguages}).
Concerning the period covered, we gather reports up to fiscal year 2022 (9.6\% of the available documents), even though the majority of the documents (54\%) refer to the fiscal year 2021 (i.e. a fiscal year is a twelve-month period that is generally equal to a calendar year).
This temporal distribution displays an expected coverage since we gathered these sustainability reports in February 2023 with the majority of the reports published throughout 2022 disclosing information concerning the previous fiscal year (2021). 

\begin{table}[htb]
    \centering 
    \small
    \caption{\textbf{Selected companies by sector.} Each sector showcases the number of companies represented. The “Companies" column shows a glimpse of the representative companies within each sector, offering a snapshot of the prominent companies chosen.}
    \begin{tabular}{c|c|l}
   \textbf{Sector} & \textbf{Number} & \textbf{Companies} \\ \hline
        Industrial	& 20 & 	Airbus, 3M Corporation, Adecco, Daikin Industries, ... \\
        Information Technology	& 16 & Apple, STMicroelectronics, Intel, NVIDIA, LG Display, ... \\

        Financials & 13	&  Bank of America, Deutsche Bank, Visa, Mastercard, ... \\

        Consumer Staples & 11 & Coca-Cola, British American Tobacco, Walmart, ... \\

        Consumer Discretionary & 11 & Amazon, Adidas, Toyota Motor, Alibaba, Tesla, ... \\

        Communication Services & 10	& Fox, Netflix, Walt Disney,  Activision Blizzard, Meta, ... \\
        
        Health Care & 10 & Amplifon, AstraZeneca, Bayer,  Johnson\&Johnson, ... \\
        
        Materials	& 9	& United States Steel, ArcelorMittal, DuPont, Croda, ... \\
        
        Energy & 9	& Saudi Aramco, TotalEnergies, Paramount Resources, ... \\

        Real Estate	& 8	& China Evergrande, Park Hotels Resorts, British Land, ... \\

        Utilities & 7 & American Electric Power, Edison, Enel, Uniper, ... \\ \hline
         & 124 & \\ 
        \hline
    \end{tabular}
    \label{tab:selectedCompaniesBySector}
\end{table}

Nevertheless, processing over six thousand sustainability reports from four thousand diverse companies poses a significant computational workload.
We consequently select a representative subset of companies by balancing sector, region, company notoriety, size and age representation.
This subset compounds to 124 companies spanning 3 continents, 15 countries and 11 GICS (Global Industry Classification Standard)\footnote{\href{https://en.wikipedia.org/wiki/Global\_Industry\_Classification\_Standard}{https://en.wikipedia.org/wiki/Global\_Industry\_Classification\_Standard}} sectors (\Cref{tab:selectedCompaniesBySector}). 
The sample covers companies established from the 19\textsuperscript{th} to the 21\textsuperscript{st} century exhibiting a wide range of market capitalization from 0.4 USD billion to 2,901 USD billion.
For this study we only considered the most recent sustainability report for each company in the selected subset, so to avoid any skewness effect. Note that the proposed methodology can be used as-is for longitudinal studies by simply using sustainability reports referred to several years.

Further details concerning the distribution of the fiscal years considered for this work, the complete list of the considered companies and the original dataset can be found in the Supplementary Material (SM) document (see Sections SM22, SM23 and SM24).

\subsubsection{ESG categorization} \label{subsubsec:esgCategorization}
We adopt the ESG categories proposed by Berg \emph{et al.} in~\cite{berg2022aggregate}.
The authors grouped, using a bottom-up methodology, nearly seven hundred ESG-related indicators from six distinct ESG data providers (Sustainalytics, S\&P Global, Refinitiv, Moody's ESG, Morgan Stanley Capital International-MSCI, and MSCI-KLD)  
into a unified categorization comprising sixty-four ESG categories. 
These categories encompass environmental, social and corporate governance issues including Employee Development, Supply Chain, Climate Risk Management, Energy, Financial Inclusion, Biodiversity, Customer Relationship, Access to Basic Services, and Board Diversity.
A complete list of all the ESG categories can be found in the Supplementary Material file (see Section SM21).

\subsubsection{ESG ratings and other company information} \label{subsubsec:companyInformation}
Rating agencies such as Refinitiv, MSCI, and Sustainalytics utilise non-financial reports and ESG-related information to systematically assess the impact of companies' activities on the environment and society. 
This assessment, typically done through numerical scores, offers stakeholders valuable measures for evaluating and comparing companies in terms of their performance related to ESG topics.

Among those, we used the Refinitiv platform,\footnote{\href{https://eikon.refinitiv.com/}{https://eikon.refinitiv.com} -- recently rebranded as LSEG Data\&Analytics} which provides high-quality financial and ESG data. 
The Refinitiv ESG ratings are given as percentage scores~\cite{refinitivMethodology}, wherein lower values (0-25) indicate poor ESG performance and a lack of transparency in publicly reporting material on ESG data (i.e., laggard companies); conversely, higher values (75-100) indicate excellent ESG performance and a high level of transparency in publicly reporting material on ESG data (i.e., leader companies).
In addition, it is worth mentioning that a zero score is assigned in the rare case that a company does not disclose any metrics or information relevant to its industry~\cite{ehlers2023deconstructing, sahin2022environmental}.

A company's ESG performance is assessed relative to industry peers due to industry-specific ESG concerns. 
For instance, packaging could be more relevant for companies producing consumer staples (e.g., the beverage industry), while materials companies (e.g., the chemicals industry) might emphasise physical-related topics such as Employee Safety.
This phenomenon has already emerged in the ESG literature~\cite{eccles2012need, khan2016corporate, busco2020preliminary} and is named ESG Industry Materiality. 
Reporting standard organizations, such as the Sustainability Accounting Standards Board (SASB),\footnote{\href{https://sasb.org}{https://sasb.org}} identify environmentally, socially or financially relevant issues for a specific industry (material factors) such as Water Management for the Non-Alcoholic Beverages industry.
This allows rating agencies to assess a company's ESG performance by outweighing the ESG topics relevant to the company's industry while helping companies focus on better disclosing these topics.

The combined ESG score is a cumulative measure of E/S/G pillars' weights, which differ by industry for the first two pillars (E and S), while the weight of the third pillar (G) remains consistent across all industries~\cite{refinitivMethodology}.

For each company, we collected twelve company features encompassing ESG scores, unchanging company details and annual financial data (regarding the same fiscal year of the considered sustainability reports).
Specifically, we gathered: the combined ESG scores, individual scores for the E/S/G pillars, company sector, industry, country, region and continent, number of employees, market capitalization, EBITDA (Earnings Before Interest, Taxes, Depreciation, and Amortization), and total liabilities.

\subsection{Data preparation}\label{subsec:dataPreparation}
\begin{figure}[htb]
    \centering
    \includegraphics[width = 0.7\textwidth]{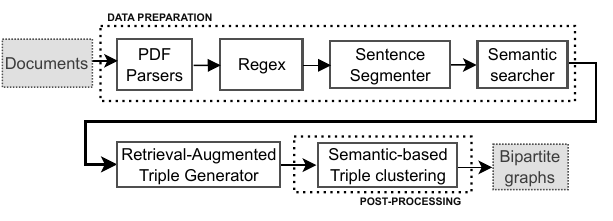}
    \caption{\textbf{Our proposed approach and its components.} 
    Given a collection of textual documents as inputs and preprocessed using different NLP methods, semantically structured insights are extracted by the retrieval-augmented triple generator.
    Bipartite graphs are created after performing a semantic-based triple clustering.}
    \label{fig:pipeline}
\end{figure}

Our NLP pipeline consists of several components (\Cref{fig:pipeline}), including some pre-processing methods, to extract structured insights from sustainability reports.
In this section, we describe the NLP methods adopted to prepare the data for our subsequent analyses. 
These pre-processing methods include extracting text from PDF files and segmenting sentences (\Cref{subsubsec:nlppipeline}) as well as using semantic search to select only sentences related to ESG topics (\Cref{subsub:semanticSearch}). 
The latter relies on the ESG categorization introduced in \Cref{subsubsec:esgCategorization}.

\subsubsection{NLP pipeline}\label{subsubsec:nlppipeline}
Sustainability reports are generally visually rich and lengthy PDF documents, for example, our sample has a median page value equal to 61 pages.
Unfortunately, the incentive of companies to present visually appealing infographics and tables, results in degraded quality of standard text extraction tools. 

Hence, for the textual part, we rely on a PDF parser (PyMuPDF,~\cite{pypiPyMuPDF})
and apply standard preprocessing steps to improve the quality of the extracted text and reduce the artefacts generated by the parser. Specifically, regular expressions 
are used to add a full stop between two sentences when missing, remove new lines in the middle of sentences and remove duplicate white spaces and lines.

Processing textual data also requires defining the granularity of the input data according to purpose, needs and limitations.
A text corpus contains textual items representing singular tokens, words, sentences, paragraphs, or entire documents.
We adopt sentence-level textual inputs as a good trade-off between the semantic meaning conveyed by a sentence 
and technical limitations (e.g., the maximum prompt length of the language model).
Consequently, after extracting textual data from the sustainability reports, we decompose each report into sentences with PySBD~\cite{sadvilkar-neumann-2020-pysbd, pypiPysbd}, a tool widely considered the state of the art in Sentence Boundary Disambiguation (SBD).

\subsubsection{Asymmetric semantic search}\label{subsub:semanticSearch}
Sustainability reports include generic and vague statements, for example, phrases such as “Air is something that surrounds us 24 hours a day”\footnote{Extrapolated from page 2 of Daikin's 2022 sustainability report.}.
Accordingly, a filtering process is required to 
consider only ESG-related sentences for downstream tasks. 
The two most well-known filtering methods for information retrieval are keyword-based and vector-based search~\cite{croft2010search}. 
We adopt the approach of neural semantic search~\cite{bast2016semantic, reimers2019sentence, muennighoff2022sgpt, guo2022semantic}, a vector-based search method, that exploits text embeddings to represent both documents and queries in the same vector space. 
Relying on text embeddings allows us to move towards a semantic-oriented filtering approach, reducing the dependency on single keywords, and thus on the ESG categorization adopted.
This representation allows one to measure the semantic similarity between a document and a query by simply computing the distance, and contrarily the similarity, between their corresponding embedded vectors~\cite{bast2016semantic}. 

Our retrieval task involves discovering sentences related to each of the ESG categories (\Cref{subsubsec:esgCategorization}) within each sentence-segmented sustainability report.
This implies working in a setting of asymmetric semantic search in which queries (ESG categories) and corpus documents (company sentences) are not interchangeable as they represent different semantic object types and have different lengths; similarly to the question answering framework~\cite{reimers2019sentence, muennighoff2022sgpt}.
In contrast, symmetric semantic search is adopted when queries and corpus documents are interchangeable such as in similar document retrieval systems~\cite{buttcher2016information, palangi2016deep, yang2019simple}.
After extensive testing, we determine that ``INSTRUCTOR-xl''~\cite{huggingfaceHkunlpinstructorxlHugging}, an instruction-tuned embedding model~\cite{su2022one}, is the most suitable choice for our specific tasks.
The authors of the model~\cite{su2022one} offer a universal instruction template (``Represent the [domain] [text type] for [task objective]:'') along with an example list. 
Since we deal with asymmetric semantic search, the instructions provided to the model vary depending on the type of input. 
For embedding the ESG categories (queries), we use the instruction ``\texttt{Represent the title for retrieving relevant statements}''. 
To embed the sentences (corpus documents), the instruction employed is ``\texttt{Represent the statement for retrieval}''.

After generating the text embeddings for a sentence-segmented sustainability report, we retrieve the most semantically relevant sentences for each ESG topic (semantic search, \Cref{tab:topKSentences}).
We set the similarity cut-off threshold $t_{sim}$ equal to 0.6 to retrieve relevant sentences to a given ESG topic.
Empirical experiments show an acceptable sentence relevance with a similarity threshold equal to or above 0.4. Thus, we adopt a cut-off point of 0.6 as a good trade-off between sentence coverage and computational workload.
In addition, we consider the top $k$ sentences (with $k = 30$) to limit the number of retrieved sentences following the aforementioned trade-off.  

This filtering layer also helps us reduce the computational time of the follow-up steps (e.g., the inference of the generative language model) and prune the resulting KG.

\begin{table}[htb]
    \centering
    \small
    \begin{tabular}{
    c|c|
    >{\centering\arraybackslash}m{10cm}
    }
         \textbf{ESG category} & \textbf{Sim score} & \textbf{Company sentence} \\ \hline
            \multirow{5}{*}{Labor Practices} 
            & 0.73 & It also guarantees \textit{the implementation of legislation on workers’ rights} defining appropriate application standards ...
            \\ \cline{2-3} 
            & 0.73 & 
            All Group workers, both in Italy and in Brazil, are covered by \textit{Collective Labour Agreements} reached with trade union organizations and ...
            \\ \cline{2-3} 
            & 0.72 & In addition to the protections and rights provided by law and \textit{the national collective labour agreement} for the 
            sector ...
            \\ \hline 
             
         \multirow{5}{*}{Waste} 
         & 0.68 & \textit{Total amount of waste by type} and disposal method 306-2 KPI A1.3 \& KPI A1.4 Environmental Management \\ \cline{2-3} 
           & 0.64 & At the same time, to \textit{avoid unnecessary material loss and waste}, the Group has formulated loss rate standards for 
           materials ... 
           \\ \cline{2-3} 
            &  0.68 & Through 
            continuous publicity and education on \textit{garbage classification}, the project wastes were gradually reduced, recycled and harmless ... 
            \\ \hline
    \end{tabular}
    \caption{\textbf{Example of the top three sentences selected for two ESG categories.} 
    The approach is capable of retrieving sentences that pertain to the two topics. However, it may also pick some meaningless sentences, such as the first one for the \texttt{Waste} category, which comes from an infographic or a tabular layout.}
    \label{tab:topKSentences}
\end{table}

\subsection{Retrieval-Augmented Triple Generation} \label{subsec:RetrievalAugmentedGeneration}
Our work aims to create a KG connecting companies, ESG topics, and actions disclosed by companies related to those topics.
To achieve this goal, we need to represent ESG-related sentences 
in a unified and standardised format: triples with a predefined semantic template.
Precisely, each ESG-oriented triple (\texttt{cat}-\texttt{pred}-\texttt{obj}) should consist of an ESG category\footnote{We use the terms “category” and “topic” interchangeably.}(\texttt{cat})
representing an ESG topic mentioned directly or indirectly in the sentence,
a predicate affecting that category (\texttt{pred}), and an entity (\texttt{obj}) related to the ESG category undergoing the predicate.
We define an action (\texttt{act}) as the concatenation between the ESG category (\texttt{cat}) and the predicate (\texttt{pred}) of a \texttt{cat-pred-obj} triple.

Consequently, our goal requires knowing the semantic meaning of words as well as defining a semantic template to generate ESG-oriented triples. 
The latest OIE techniques (\Cref{subsec:rwKgGeneration}) incorporate semantic information for extracting structured information, yet they rely only on the syntactical structure of the sentence.
For example, given the sentence: 
``\emph{Microsoft has invested 125 million in cutting-edge recycling technologies}''\footnote{This sentence is created for explanatory purposes, not reflecting real information.},
conventional OIE techniques \cite{oie_demo} would identify and generate a traditional SPO triple as the following: \texttt{(Microsoft, Invested, 125 million)}.
Although the above SPO triple can well represent the semantic meaning conveyed, it would not be suitable for our goal.
Indeed, the ideal triple would have been: \texttt{(Waste, Investment in, Cutting-edge recycling technologies)}.
Generating the latter requires defining a semantically-aware triple template.
Firstly, the entity \texttt{Waste}, representing an ESG category, is not explicitly mentioned, although it could be inferred from the term \texttt{recycling technologies}.
This type of inference jointly involves information extraction and semantic classification tasks.
Secondly, our goal requires extracting ESG-related actions rather than generic statements. Hence, triples should envelop predicates and objects related to an ESG category.
For instance, given that ideal triple, the action (\texttt{act}) is defined as the ESG category \texttt{Waste} concatenated with the predicate \texttt{Investment in}, resulting in the action “\texttt{Waste:Investment in}".

LLMs have already demonstrated abilities in semantic understanding and handling a broad range of NLP-related tasks~\cite{brown2020language, reynolds2021prompt}. 
Accordingly, in this work, we employ instruction-tuned LLMs, the In-Context Learning technique and the prominent RAG paradigm~\cite{lewis2020retrieval} to address this challenge. 
Our work exploits these techniques to provide an LLM with an input (ESG-related sentence) and an external context (input-output examples and a semantic schema) to extract structured information from the sentence.

\begin{figure}[htp]
    \centering
    \includegraphics[width = 0.75\textwidth]{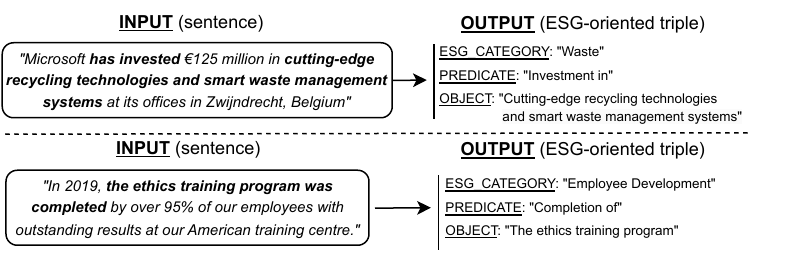}
    \caption{
    \textbf{Two labelled examples included within our model instruction.} The input sentences were created to cover different syntactical structures, and do not represent actual information.}
    \label{fig:labelledExamples}
\end{figure}

We choose the {\texttt{Kor}} library~\cite{KorPyPI} to create in-context instructions for LLMs. 
{\texttt{Kor}} allows to programmatically construct prompts by specifying the semantic data schema for the ideal triples (\texttt{cat}-\texttt{pred}-\texttt{obj}) as well as including labelled examples. 
A labelled example connects an input sentence with the desired output, an ESG-oriented triple.
\Cref{fig:labelledExamples} exhibits two labelled examples included in the model instruction to leverage the In-Context Learning abilities of the LLM.
In the model instruction, each element of the triple (\texttt{cat, pred} and \texttt{obj}) is declared with a unique name and a natural language description conveying its semantic meaning.
The \texttt{Kor} library then uses this information to generate, by means of a predefined template (\Cref{fig:promptTempate}), a textual instruction to prompt an instruction-tuned LLM.
We integrated the sixty-four ESG categories of the ESG categorization (\Cref{subsubsec:esgCategorization}) in the description of the attribute \texttt{cat}. 
The LLM leverages this list of ESG topics as semantic guidance, aiding itself in the generation of the ESG category for a given input sentence. 
It achieves this by mimicking the results of a supervised text classifier, whose labels are those of the ESG categorization, and extrapolating those labels with semantic generalization.
The full model instruction is exhibited in Section SM2 of the Supplementary Material (SM) document.

\begin{figure}[htb]
\begin{minipage}{0.9\textwidth}
\begin{quotation}{ \scriptsize \ttfamily \hspace{-6.5mm}
Your goal is to extract structured information from the user's input that matches the form described below. When extracting information please make sure it matches the type of information exactly. 
Do not add any attributes that do not appear in the schema shown below. 
\newline 
\texttt{\$DATASCHEMA} \newline 
Please output the extracted information in JSON format. Do not output anything except for the extracted information. Do not add any clarifying information. Do not add any fields that are not in the schema. If the text contains attributes that do not appear in the schema, please ignore them. All output must be in JSON format and follow the schema specified above. Wrap the JSON in $<$json$>$ tags. \newline 
\texttt{\$EXAMPLES} \newline 
input: \texttt{\$INPUT} \newline
output:
}\end{quotation}
\end{minipage}
\caption{\textbf{The instruction template used to prompt the generative LLM}. The instruction template is created and compiled by the \texttt{Kor} library to prompt an LLM to extract structured data using In-Context Learning. \texttt{DATASCHEMA} is a placeholder for the output data schema. \texttt{EXAMPLES} is a placeholder for the labelled examples in the format of input-output pairs. While \texttt{INPUT} represents an ESG-related sentence from which structured data needs to be retrieved.}
\label{fig:promptTempate}
\end{figure}

We tested different instruction-tuned LLMs such as Google's Flan-T5~\cite{chung2022scaling} and LLaMA-based models~\cite{touvron2023llama}.
We empirically found that LLaMA-based models (e.g., Alpaca~\cite{taori2023stanford}) generate better results when prompted to extract structured information, with \texttt{WizardLM-7B}~\cite{xu2023wizardlm, huggingfaceTheBlokewizardLM7BHFHugging} producing the highest-quality results. 
\Cref{apx:LLMParameters} and \Cref{apx:experimentsLLMs} showcase additional information regarding empirical experiments conducted on different LLMs, as well as the selection of specific hyperparameters for the LLM generation process.

\subsection{Knowledge Graph generation}\label{subsec:knowledgeGeneration}
Before constructing a KG using the generated triples \texttt{cat}-\texttt{pred}-\texttt{obj}, we apply a data-cleaning process to reduce data redundancy.
The redundancy in the KG comes from nodes and edges representing similar concepts and their relationships multiple times.

\begin{figure}[b]
    \centering
    \includegraphics[width = 0.5\textwidth]{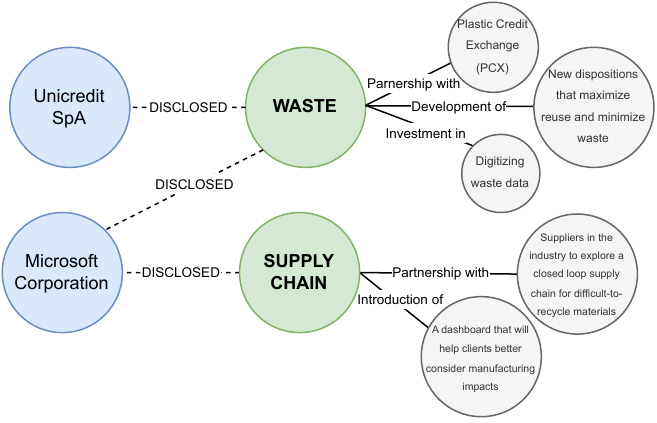}
    \caption{
    \textbf{An example of a portion of the Knowledge Graph generated using our methodology.} It portrays the ESG-oriented triples generated using our approach. Blue nodes represent company nodes which are connected to the ESG categories (green nodes) disclosed in the companies' sustainability reports. Category nodes (\texttt{cat}) are connected via a labelled edge (\texttt{pred}) to the predicate object (\texttt{obj}, grey nodes).}
    \label{fig:portionKG}
\end{figure}

To achieve this goal, we perform semantic clustering on all the ESG categories (\texttt{cat}) and predicates (\texttt{pred}) included in the generated triples.
Firstly, we generate text embeddings using the ``INSTRUCTOR-xl'' embedding model~\cite{su2022one} with the model instruction ``\texttt{Represent the title}''.  
Secondly, semantic clusters are discovered as high-density regions in the embedded vector space using cosine similarity as a metric. 
We conduct several empirical experiments to evaluate the cluster goodness using different similarity cut-off thresholds, ranging from 0.5 to 0.9.
Eventually, we adopt a similarity cut-off point of 0.8, as it strikes a good balance between the semantic coherence of the cluster elements (cluster quality) and the cluster sizes.
Finally, we label each cluster with its centroid and use cluster labels to replace the original ESG categories and predicates of the original triples.
For instance, the predicate cluster labelled \texttt{Partnership with} groups 103 different predicates encompassing \texttt{Working together with, Partnering with others to} and \texttt{Collaborating of}. 
Other cluster examples and replacing examples are reported in the Supplementary Material file (Section SM1). 
\Cref{fig:portionKG} exhibits, for explanatory purposes, a portion of the KG generated using our methodology.


\subsection{Approaches for statistical analyses} \label{subsec:analysisMethods}
In the Results Section (\Cref{sec:results}), we mostly deal with undirected bipartite graphs obtained from the original KG.
A bipartite graph is a graph whose vertices can be divided into two distinct and independent sets or partitions~\cite{asratian1998bipartite, estrada2012structure,newman2010network}. 
It can be described through its binary \emph{bi-adjacency} matrix $\mathbf{B}$, a $\{0,1\}^{n \times m}$ matrix where $n$ and $m$ are the numbers of nodes in the two partitions.
A bipartite graph can consequently be seen as a special type of knowledge graph whose nodes can be divided into two distinct and independent partitions. 
Its graph edges are accordingly context-dependent and change based on the perspective used to generate the bipartite graph.

Specifically, we create three distinct bipartite graphs for the analyses of our findings through node and edge filtering of the comprehensive KG; though isolating distinct types of nodes, and their relative connections, from the original graph.
The creation of different two-fold representations (bipartite graphs) can help conduct downstream analyses on specific relationships among different types of nodes included in the comprehensive knowledge graph.
This allows us to analyse the extracted insights using three distinct perspectives: 
\begin{enumerate}
    \item the predicates (pred) disclosed with each ESG category (cat): analysed using the category-predicate bipartite graph $\mathbb{B}_{\textrm{catpred}}$;
    \item the ESG categories (cat) disclosed by each company: analysed using the company-category bipartite graph $\mathbb{B}_{\textrm{cocat}}$;
    \item the actions (act) disclosed by each company: analysed using the company-action bipartite graph $\mathbb{B}_{\textrm{coact}}$.
\end{enumerate}

A table encompassing the number of partition nodes, the number of edges and the density for each bipartite graph is provided in the Supplementary Material (see Section SM4).

\subsubsection{Bipartite graph statistics}
Most of the unipartite graph metrics can be extended to the bipartite case~\cite{asratian1998bipartite,newman2010network}.
Specifically, here we compute the bipartite variants of network statistics such as degree centrality, closeness centrality and betweenness centrality~\cite{newman2010network, brandes2005network}.

The degree centrality of a partition node is the fraction of the nodes of the other partition connected to it~\cite{zhang2017degree}.
The closeness centrality of a node~\cite{estrada2012structure, newman2010network} is determined by calculating its average shortest path distance to all other nodes and represents the efficiency of a node to be connected directly to nodes from the other partition and indirectly to nodes from the same partition~\cite{faust1997centrality}.
For instance given $\mathbb{B}_{\textrm{cocat}}$, a company node with a high closeness score indicates the company is connected, and thus it is close, to many category nodes which in turn are connected to several other company nodes.
Lastly, betweenness centrality~\cite{estrada2012structure, barthelemy2004betweenness} assesses the level of influence a node holds over information flow within a graph. 
In the context of bipartite graphs, it identifies nodes serving as critical mediators in enabling interactions between the two separate node partitions~\cite{asratian1998bipartite}.



\subsubsection{ESG-related actions' variability} \label{subsub:actionVariability} 
We leverage information theory to assess the entropy of ESG topics based on 
their associated predicates yielded from companies' sustainability reports.
Specifically, we adopt Shannon's entropy (\Cref{eq:entropy},~\cite{cover1999elements}) to measure the
information, and thus the variability, present in a set of events $\mathcal{X}$ through their respective occurrence probabilities $p(x)$:
\begin{equation}
    H( \mathcal{X}) := - \sum_{x \in \mathcal{X}} p(x) \ln  p(x) 
    \label{eq:entropy}
\end{equation}

In our context, the events $\mathcal{X}$ are the predicates disclosed by all companies for a given ESG topic, while $p(x)$ represents their relative occurrences.
High entropy denotes high variability in the predicate occurrences, indicating an ESG topic addressed through many actions (predicates) with an almost uniform probability of predicate occurrence. 
On the other hand, low entropy indicates the predominance of a limited set of predicates for an ESG topic.

\subsubsection{Similarity analysis} \label{subsub:simAnalysisMethod}
We estimate company similarities based on jointly disclosed ESG-related actions through the Jaccard similarity coefficient~\cite{costa2021further}, which measures the similarity between two sets as the cardinality of their intersection over the cardinality of their union:
\begin{equation}
    J(\mathcal{A}_{c_1}, \mathcal{A}_{c_2}) = \frac{|\mathcal{A}_{c_1} \cap \mathcal{A}_{c_2}|}{|\mathcal{A}_{c_1} \cup \mathcal{A}_{c_2}|} \\
    \label{eq:jaccardSim}
\end{equation} 
where $\mathcal{A}_{c_i}$ is the set of ESG-related actions disclosed by company $c_i$.
To mitigate the influence of stochastic fluctuations on the similarity score, we generated a null model~\cite{gotelli2012statistical} with a bootstrapping technique by computing company similarities on the randomised action sets through 1,000 simulations, and substracted this null model from the observed company similarities. 


\subsubsection{Correlation analysis}
We evaluate whether company similarities in terms of jointly disclosed ESG-related actions (\Cref{subsub:simAnalysisMethod}) are correlated to similarities in ESG scores or other company characteristics such as market capitalization or geographical location.
We first measure feature similarities through different strategies, ensuring the same numerical range and monotonicity. 
The similarities in numerical features, such as ESG scores, are measured by computing the absolute numerical difference normalised using max-min scaling~\cite{zheng2018feature}.
While, similarities in textual features, such as company sectors, are first embedded using the ``INSTRUCTOR-xl'' embedding model~\cite{su2022one}, and then their semantic similarities are assessed through the cosine similarity normalised in [0,1] using min-max scaling~\cite{zheng2018feature}.
A complete list of all the features and measures used can be found in the Supplementary Material (see Section SM9). 

Afterwards, we perform a bivariate analysis through a correlation analysis to measure the monotonic association between action similarities and similarities of other company features. 
We rely on Kendall's $\tau$ correlation coefficient (\Cref{eq:corrCoeff},~\cite{abdi2007kendall}), a non-parametric and rank-based statistic computed as:
\begin{equation}\label{eq:corrCoeff}
\tau = \frac{n_c - n_d}{n_c + n_d} 
\end{equation}
where $n_c$ and $n_d$ are the numbers of concordant and discordant pairs respectively. 
Rank-based correlation methods overcome some limitations of traditional correlation methods such as the well-known Pearson correlation coefficient~\cite{cohen2009pearson}: they can measure nonlinear monotonic relationships, are more robust against outliers and normality assumption is not required~\cite{ott2015introduction}.
High positive coefficients express a high level of order consistency in the company similarities sorted according to actions' and other similarities, while high negative coefficients occur when these two similarities are sorted reversely~\cite{abdi2007kendall}.

\subsubsection{Interpretability of ESG scores}\label{subsubsec:ESGinterpretability}
Lastly, we investigate the interpretability of ESG scores through linear regression and the SHAP (SHapley Additive exPlanations) framework~\cite{NIPS2017_7062}.
Here, we investigate the most impacting factors on the ESG scores of companies by exploiting the interpretability of a first-order linear regression model. 

The model predictors are based on our findings and other company information (\Cref{subsubsec:companyInformation}).
We first use as predictors the percentage of the top ten most disclosed ESG categories for each company.
For example, if a company has ten percent of its generated triples concerning the ESG category \texttt{Waste}, and that is within its top ten most disclosed topics, 
the feature \texttt{Category:Waste} for this observation has a value of 0.1.
We also consider the proportion of the E/S/G pillars based on all the disclosed categories for each company. 
In addition, we compute the category and action entropy for each company, indirectly representing the cardinality of the ESG categories and actions disclosed in its sustainability report (\Cref{subsub:actionVariability}).
Lastly, we consider nine company-related features as further predictors encompassing five company characteristics and four annual financial attributes.
Specifically, we include in our predictors the company sector, the country, region and continent of its headquarters, and its incorporation year. 
On the other hand, we also include financial features concerning the fiscal year of the analysed sustainability report: EBITDA (Earnings Before Interest, Taxes, Depreciation and Amortisation), liabilities, market capitalization and the number of employees.

\begin{figure}[htp]
    \centering
    \includegraphics[width = 0.8\textwidth]{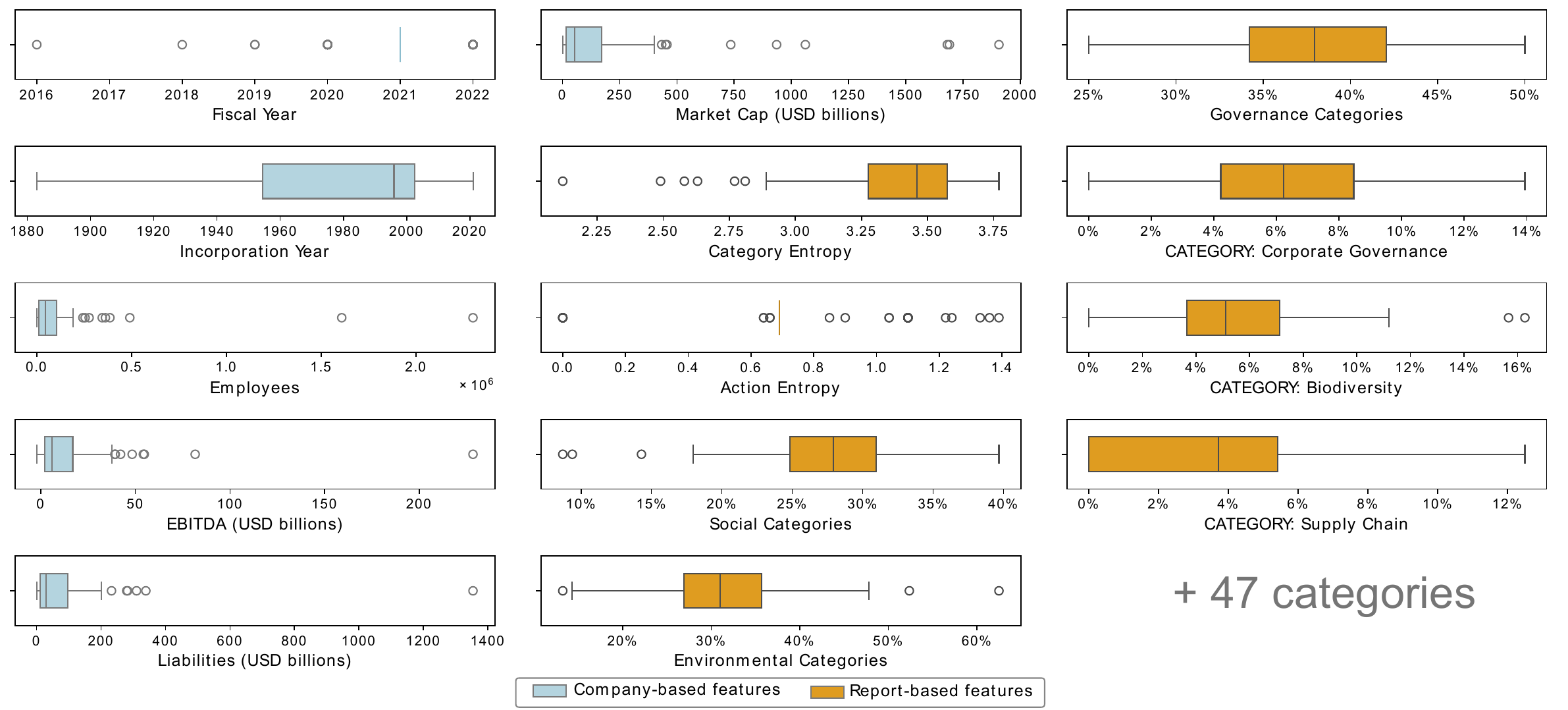}
    \caption{\textbf{Descriptive statistics of the numerical features.} They are used as part of the predictors of the Ordinary Least Squares (OLS) model.
    The features labelled with the starting word “category" reflect the percentage disclosure of an ESG topic.
    To enhance readability, the figure exhibits only three category features for explanatory purposes. The statistics are presented before standardisation.}
    \label{fig:numericalPredictors}
\end{figure}

The descriptive statistics of these report-based and company-based predictors are exhibited in \Cref{fig:numericalPredictors} and \Cref{tab:categoricalPredictors}. 
Categorical variables, such as the company sector, are turned into binary indicator variables (i.e., dummy variables~\cite{draper1998applied}),  generating a total of 97 features whereas standardisation is applied to those numerical such as market capitalization.
An example of an observation, encompassing a complete list of features, is exhibited in Section SM14 of the Supplementary Material (SM) document.

\begin{table}[htp]
    \centering
    \small
    \caption{\textbf{Descriptive statistics of the categorical variables}. They are used as part of the predictors of the Ordinary Least Squares (OLS) model. All these variables are transformed into binary indicator variables for each observation.}
    \begin{tabular}{
    >{\centering\arraybackslash}m{2cm}|
    >{\centering\arraybackslash}m{1.1cm}|
    >{\centering\arraybackslash}m{10.5cm}
    }
        \textbf{Feature name} & \textbf{Unique} & \textbf{Top frequent categories (\%)} \\ \hline
        Region & 3 & Americas (56.3\%), Europe (29.9 \%), Asia (13.8\%)\\
        Subregion & 7 & Norther America (55.2\%), Western Europe (13.8\%), Eastern Asia (12.6\%), ...  \\
        Country & 15 & United States of America (51.7\%), Italy (8\%), United Kingdom (6.9\%), ... \\
        Sector & 11 & Industrials (14.9\%), Information Technology (11.5\%), ... \\ \hline 
    \end{tabular}
    \label{tab:categoricalPredictors}
\end{table}

Then, we adopt an Ordinary Least Squares (OLS,~\cite{dismuke2006ordinary}) regression with Elastic Net Regularization~\cite{zou2005regularization} to perform inference on companies' ESG scores available.
We adopt Elastic Net Regularization, a generalisation of the LASSO method~\cite{hastie2009elements}, to perform both feature selection and training regularisation.
We choose this regularisation method as it improves performance when the number of predictors ($\vert features \vert = 97$) is higher than the observations ($\vert companies \vert = 89$) as well as in the presence of strong pairwise correlations~\cite{zou2005regularization}. 
We train the OLS model using the Elastic Net cost function and an eight-fold cross-validation approach~\cite{berrar2019cross} (see Section SM16).
The performance of the OLS model is discussed in \Cref{apx:OLSPerfomance}.

Subsequently, we employ the SHAP framework~\cite{NIPS2017_7062} to investigate which predictors impact the inference of ESG scores most.
SHAP is a model-agnostic and additive feature importance measure which is based on cooperative game theory. 
It provides local interpretations of model predictions as additive sums of the directed effects of each predictor using a conditional expectation function.
SHAP starts from the prior knowledge of the expected model output $E[f(X)]$, and then evaluates, for each model prediction, the magnitude and direction changes in this expected value (SHAP values) when conditioned on each predictor.
It thus quantifies the magnitude and direction of the observed effects of each predictor. 
Predictors with positive SHAP values affect the expected model output with additive increments, conversely, those with negative values have additive decrement impacts.


\section{Results} \label{sec:results}
In this section, we first report the network statistics computed at the node level from the three bipartite graphs in \Cref{subsec:bipartiteAnalysis}.
Next, a diversity analysis examines how ESG topics are disclosed across companies and different sectors (\Cref{subsec:diversityAnalysis}).
\Cref{subsec:companySimilarity} reports company similarities according to jointly disclosed ESG actions. 
The follow-up section (\ref{subsec:correlationAnalysis}) addresses whether these company similarities are associated with similarities in other company information.
Finally, we evaluate the interpretability of ESG scores by investigating the most impacting factual aspects (\Cref{subsec:ESGinterpretability}).
A qualitative analysis of the generated triples and an ablation study on the model instruction are exhibited in \Cref{apx:qualitativeAnalysis} and \Cref{apx:ablationStudy}.


\subsection{Bipartite graphs' analysis}\label{subsec:bipartiteAnalysis}
We here present some network statistics concerning the three bipartite graphs 
$\mathbb{B}_{\textrm{cocat}}$, $\mathbb{B}_{\textrm{catpred}}$, and $\mathbb{B}_{\textrm{coact}}$.
Further statistics and extensive tables for all three bipartite graphs are shown in the Supplementary Material document in Sections SM4, SM5 and SM6.

\paragraph{\textbf{The ESG movement encompasses many socially responsible issues.}}
Our data-driven methodology unveils that the one hundred and twenty-four companies disclosed 542 distinct ESG topics/categories in their sustainability reports. 
The company-category bipartite graph $\mathbb{B}_{\textrm{cocat}}$ has an average degree distribution equal to almost 11\%, making this graph relatively connected.
There are however some mainstream ESG categories: \texttt{Climate Risk Management, Supply Chain, Energy} and \texttt{Corporate Governance} are connected to, and thus disclosed by, almost all the considered companies (degree $>$ 92\%, \Cref{tab:categoryNodesMetrics}). 
Conversely, the list of the least disclosed categories is rather surprising: \texttt{Market Responsibility}, \texttt{Anti-Discrimination} and \texttt{LGBTQ+ Inclusion} are connected to, and thus disclosed by, less than 5\% of all the considered companies (\Cref{tab:categoryNodesMetrics}).

\begin{table}[htp]
    \centering
    \caption{\textbf{Graph metrics} of a sample of all the 542 category nodes of the bipartite graph  $\mathbb{B}_{\textrm{cocat}}$.}
    \begin{tabular}{c|c|c|c}
    \textbf{Category node} (\texttt{cat}) & \textbf{Degree} (\%) & \textbf{Closeness} (\%) & \textbf{Betweenness} (\%) \\ \hline
    Supply Chain & 95.3 & 98.9	& 2.3 \\
    Climate Risk Management	& 94.5 & 98.5	& 2.3 \\ 
    Corporate Governance & 93.0	& 98.1	& 2.1 \\
    Community and Society & 91.4 & 98.1	& 2.1 \\
    Philanthropy & 88.3 & 96.8	& 2.0 \\
    Packaging & 50.0 & 77.7	& 0.6 \\
    Human Resources	& 39.8 & 74.2 & 0.4 \\
    LGBTQ+ Inclusion & 4.7 & 55.4 &	0.0 \\
    Anti-Discrimination	& 2.3 & 52.1 & 0.0 \\
    Marketing Responsibly & 0.8 & 47.8 & 0.0 \\ \hline
    \end{tabular}
    \label{tab:categoryNodesMetrics}
\end{table}

\paragraph{\textbf{ESG topics are addressed from several perspectives, with some frequent ones.}} 
The average degree centrality of the category-predicate bipartite graph $\mathbb{B}_{\textrm{catpred}}$ is less than 1\%. 
There are however some predominant predicate nodes (see Section SM6) that are associated with more than ninety ESG categories (degree $\geq$ 16.6\%) such as \texttt{Commitment and involvement with} (113 categories), \texttt{Advisor support for} (102), \texttt{Partnership with} (97), and \texttt{Establishment of} (94).
These prominent nodes interestingly exhibit high closeness centrality ($>$ 88\%) in contrast to their relatively low degree centrality values. 
This indicates that common category nodes indirectly connect them.

\paragraph{\textbf{Common actions are the exception.}}
The company-action bipartite graph $\mathbb{B}_{\textrm{coact}}$ connects the company nodes to almost twenty thousand different ESG-related actions (19,574) disclosed in companies' sustainability reports. 
However, there are a few prominent actions disclosed by the majority of the considered companies:
\texttt{AIR EMISSION: Reduction of} (degree of, and thus connected to, the 70\% of the companies), \texttt{ENERGY:Reduction of} (61\%), \texttt{PHILANTHROPY:Donation by} (60\%) and \texttt{CLIMATE RISK MANAGEMENT:Assessment of} (56\%).


\subsection{Diversity analysis on disclosing ESG categories}\label{subsec:diversityAnalysis}
In this section, we analyse the differences in disclosing and addressing ESG topics in companies' sustainability reports.


\paragraph{\textbf{Companies approach ESG topics from plenty of perspectives, especially those generic and vague.}}
The predicates associated with each ESG category vary significantly with an average Shannon's entropy of 1.5 nats.
This broad action variety is predominant in generic and umbrella ESG topics such as \texttt{Corporate Governance, Human Rights}, and \texttt{Supply Chain} (\Cref{tab:entropyCatPred}). 
For example, when addressing \texttt{Product Safety}, companies approach it from different perspectives, ranging from developments (2.3\%) to regulatory compliance (2.3\%) and assessments (4.4\%, \Cref{tab:entropyCatPred}).
However, a notable correlation ($corr = 0.84$) exists between the entropy of categories and the number of companies disclosing information about a particular ESG category. This suggests differences in how each company approaches these ESG topics.
\begin{table}[htp]
    \centering
    \small
    \caption{\textbf{A sample of the ESG categories with their entropy values computed.} 
     The three most frequent category predicates are reported alongside the percentage of companies disclosing that category.}
    \begin{tabular}{
    >{\centering\arraybackslash}m{3.1cm}|
    >{\centering\arraybackslash}m{1.3cm}| 
    >{\centering\arraybackslash}m{2cm}| 
    >{\centering\arraybackslash}m{8cm}}
        \textbf{ESG category} (\texttt{cat}) & \textbf{Entropy} (nats) & \textbf{Companies} (\%) & \textbf{Category predicates} (\texttt{pred})  \\ \hline
        Supply Chain & 5.72 & 95 & Partnership with (3.7\%), Assessment of (2.4\%), Engagement in (2.2\%)\\ \hline
        Corporate Governance & 5.62 & 94 &  Establishment of (4.3\%), Commitment and involvement with (3.7\%), Overseeing (2.9\%) \\\hline
        Human Rights & 5.62 & 90 &  Commitment and involvement with (5.1\%), Respect for (2.6\%), Assessment of (2.4\%)\\\hline
        Product Safety & 5.27 & 69 & Assessment of (4.4\%), Compliance with (2.3\%), Development and implementation of (2.3\%) \\ \hline
        Food Waste & 2.44 & 6 &Commitment and involvement with (14.3\%), Reduction of (14.3\%), Development of (7.1\%) \\ \hline
        Anti-Slavery Practices & 1.61 & 4 & Undertaking of (20\%), Integration of (20\%), Required training (20\%) \\ \hline
    \end{tabular}
    \label{tab:entropyCatPred}
\end{table}

\paragraph{\textbf{Cross-sector vs sector-focused topics.}}
A proportion of almost 12\% among all the 542 ESG categories is disclosed across all company sectors, encompassing various umbrella aspects such as \texttt{Climate Risk Management}, \texttt{Supply Chain} and \texttt{Business Ethics}. 
However, it is noteworthy that certain sectors emphasise specific topics more than others (\Cref{tab:categoryStats}).
For example, \texttt{Packaging} is more emphasised in Consumer Staples companies, such as PepsiCo (18\% of all the company triples), Coca-Cola (9\%), Monster Beverage (6\%), and Tesco (5\%). 
This category accounts for 4.9\% of all generated company triples within this sector as exhibited in  \Cref{tab:categoryStats}. 
Another example is \texttt{Water} which is more stressed by companies operating in water-intensive sectors such as Consumer Staples (6.6\%, \Cref{tab:categoryStats}, e.g., Coca-Cola) and Materials (4.9\%, e.g., DuPont) rather than Financials (0.5\%, e.g., Goldman Sachs).


\begin{table}[htb]
    \centering
    \small
    \caption{\textbf{Sample of all the ESG categories disclosed by companies in their sustainability reports.} This table exhibits the category coverage through different percentages concerning: (i) the company triples including a category, (ii) the companies reporting a category (iii) also aggregated by sector, and (iv) the company triples aggregated by sector.}
    \begin{tabular}{
    >{\centering\arraybackslash}m{3cm}|
    >{\centering\arraybackslash}m{1cm}|
    >{\centering\arraybackslash}m{1.7cm}|
    >{\centering\arraybackslash}m{1.2cm}|
    >{\centering\arraybackslash}m{6.1cm}
    }
   \textbf{ESG category} (\texttt{cat}) & \textbf{Triples} (\%) & \textbf{Companies} (\%) & \textbf{Sectors} (\%)  & \textbf{Triples per company sector} (\%) \\ \hline
    Corporate Governance & 6.7 & 94 & 100 & Industrials (7.8\%), Financial Services (7.5\%), Healthcare (7.4\%) \\ \hline
    Air Emissions & 3.7	& 92 & 100 & Energy (6.9\%), Basic Materials (4\%), Industrials (3.8\%) \\ \hline
    Water & 3.1	& 85 & 100 & Consumer Defensive (6.6\%), Basic Materials (4.9\%), Energy (4.4\%) \\ \hline
    Green Buildings & 1.5 & 88 & 100 & Real Estate (3.5\%), Consumer Cyclical (1.8\%), Technology (1.6\%) \\ \hline

    Packaging & 0.9 & 49 & 100 & Consumer Defensive (4.9\%), Consumer Cyclical (1.5\%), Technology (0.9\%)\\ \hline
    Business Ethics	& 0.3 & 61 & 100 & Healthcare (0.7\%), Utilities (0.5\%), Industrials (0.4\%), \\ \hline
    \end{tabular}
    \label{tab:categoryStats}
\end{table}

\paragraph{\textbf{Comparison with ESG materiality at the sector and industry level}}
We compare the data-driven evidence of sector-focused ESG topics with the sector-level ESG materiality identified by Khan \emph{et al.}~\cite{khan2016corporate}.
We explore the relevant issues (materiality) identified for the Financials sector for explanatory purposes as is one of the common sectors between the authors' work and ours.
There are thirteen financial companies in our sample: nine commercial banks (e.g., Deutsche Bank), two credit service companies (e.g., Mastercard), one insurance company (Assicurazioni Generali) and one asset management company (3i Group).
Our data-driven findings unveil that the companies' sustainability reports address the ten issues identified as relevant for this sector with different importance.
Financial companies, for example, extensively disclosed actions concerning:
\textit{environmental, and social impacts on core assets and operations} (\texttt{Climate Risk Management}: 7\% of all the sector triples),
\textit{Diversity and Inclusion} (\texttt{Financial Inclusion}: 6.1\% and \texttt{Board Diversity}: 5.2\%) and 
\textit{Lifecyle impacts of products and services} (\texttt{Product Sustainability}: 4.1\%).
On the other hand, companies' non-financial disclosures neglect to address some relevant issues encompassing: 
\textit{Access and affordability} (\texttt{Access to Basic Services}: 0.8\%, 
\texttt{Accessibility}: 0.3\%, and \texttt{Access to Information}: 0.1\%),
\textit{Fair marketing and advertising} (\texttt{Marketing} and \texttt{Advertising}: 0.1\%), and
\textit{Business ethics and transparency of payments} (\texttt{Business Ethics}: 0.2\% and \texttt{Anti-Money Laundering}: 0.1\%).
A table exhibiting the comparison with all ten issues can be found in Section SM10 in the Supplementary Material (SM) document.

Furthermore, we select UniCredit, a financial company operating in the industry of Commercial Banks, as an explanatory example to compare our findings with the relevant disclosure topics outlined in SASB standards for its industry. 
This reporting standard organization identifies six important disclosure topics for commercial banks~\cite{FindIndustryTopics} which were addressed differently in the company's sustainability report.
The company focused on industry-relevant issues encompassing:
\textit{Financial Inclusion \& Capacity Building} (\texttt{Financial Inclusion}: 4.6\%) and \textit{Data Security} (\texttt{Data Privacy}: 2.4\%).
It however neglected to disclosure much information concerning issues such as \textit{Financed Emissions} (\texttt{Social Responsibility}: 0.5\%) and \textit{Incorporation of Environmental, Social, and Governance Factors in Credit Analysis} (\texttt{Environmental Risk Assessment}: 0.4\%).

\paragraph{\textbf{Different sectors employ tailored actions to address ESG topics.}}
Although there are a few widely disclosed actions (\Cref{subsec:bipartiteAnalysis}), the same action is disclosed, on average, by less than 2\% of the considered companies.
Whereas, only 15\% of the company sectors are, on average, engaged in the same action. 
This unveils different priorities and a variety of approaches among companies and sectors.
For example, the \texttt{Assessment of} aspects concerning \texttt{Climate Risk Management} are more emphasised by Real Estate companies which manage assets vulnerable to climate risks, such as Park Hotels Resorts (2\% of all the company triples) and Sun Communities (1\%).
Conversely, Materials companies such as United States Steel (1\%), Yamana Gold (1\%) and Aluminum Corporation of China (1\%), emphasise instead the \texttt{Commitment and involvement} concerning \texttt{Employee Safety}, a worker-related topic.
Further extensive tables can be found in the Supplementary Material document (see Section SM3).

\subsection{Company similarities based on disclosed ESG-related actions}  \label{subsec:companySimilarity}
Here, we discuss company similarities according to jointly disclosed actions using the Jaccard similarity coefficient (see \Cref{subsec:analysisMethods}). 

\begin{figure}[htp]
    \centering
    \includegraphics[width = 0.85\textwidth]{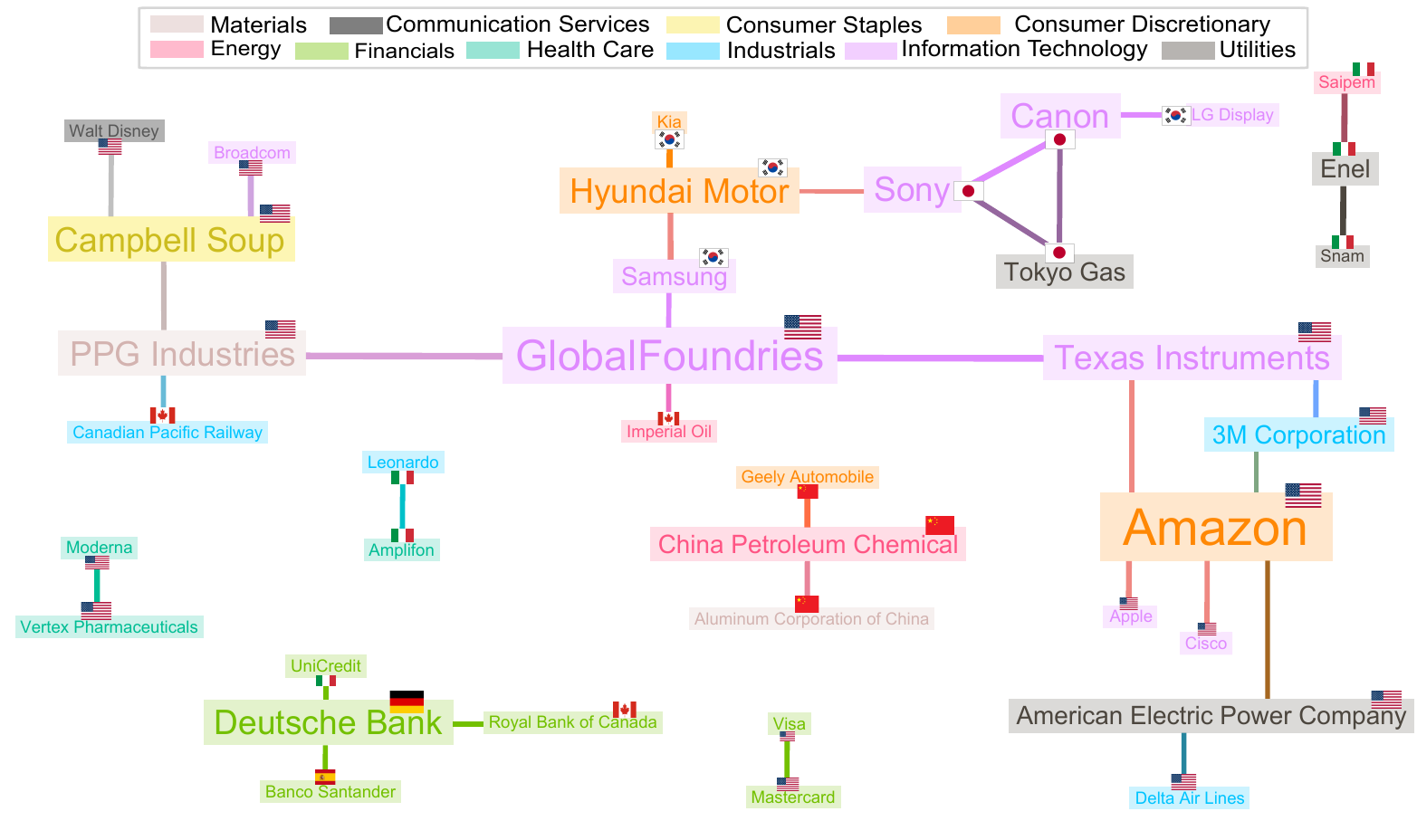}
    \caption{\textbf{A network diagram linking companies that report similar actions, determined by the Jaccard similarity coefficient.} 
    It exhibits only connections between companies with a similarity equal to or greater than 6\%.
    Node colour corresponds to distinct sectors, and node size is proportional to their connectivity. 
    Some connections are noteworthy for linking companies within the same sectors or geographical regions.}
    \label{fig:jaccardSimilarity}
\end{figure}

\paragraph{\textbf{Companies from the same sectors tend to perform similar actions.}}
For example, as depicted in \Cref{fig:jaccardSimilarity} and outlined in \Cref{tab:jaccardSimilarities}, five companies among the top ten similar companies of \texttt{Deutsche Bank} are banks too: \texttt{Royal Bank of Canada} (action similarity equal to 7\%), \texttt{Banco Santander} (6\%) and \texttt{UniCredit} (6\%). 
Notably, \texttt{Visa} and \texttt{Mastercard}, both operating in the Credit Services industry, form a distinct group (\Cref{fig:jaccardSimilarity}).
Comparably, action similarities emerge in companies operating in the healthcare sector: Moderna, Vertex Pharmaceuticals and AstraZeneca (\Cref{tab:jaccardSimilarities}).
Further details are visible in Section SM8 of the Supplementary Material document.

\paragraph{\textbf{Companies from the same geographical region tend to perform similar actions.}}

For example, as can be visually noted in \Cref{fig:jaccardSimilarity}, 80\% of the ten most similar companies of \texttt{Sony} (Japan, Eastern Asia) are companies from the same geographical region: 40\% from Japan and 40\% from South Korea. 
Similarly, \texttt{Geely Automobile} (China, Eastern Asia) has 70\% of its ten most similar companies from the same region too: 40\% from China and 30\% from South Korea. 
On the west side, there are six European companies in the ten most similar companies of \texttt{Enel} (Italy, Southern Europe) with Italian companies representing 40\% of the total.

\begin{table}[htp]
    \centering
    \small
    \caption{\textbf{A company sample with the top three most reported actions and the most similar companies for each.} Company similarities are assessed by computing the Jaccard similarity on the companies' disclosed action set.}
    \label{tab:jaccardSimilarities}
    \hspace*{-4mm}
    \begin{tabular}{
    >{\centering\arraybackslash}m{1.55cm}|
    c|
    >{\centering\arraybackslash}m{5.5cm}}
        \textbf{Company} & \textbf{Most reported actions} & \textbf{Most similar companies} \\ \hline
        Sony & \makecell{ 
        PACKAGING: Reduction of (x7) \\ 
        PHILANTHROPY: Advisory support for (x6) \\
        CORPORATE GOVERNANCE: Establishment of (x6) }
        & Canon (7\%), Tokyo Gas (6\%), Hyundai Motor (6\%),  Toshiba (6\%), Kia (6\%) 
        \\ \hline
        
        Deutsche Bank &\makecell{
        ENERGY: Reduction of (x4) \\
        BIODIVERSITY: Promotion of (x4) \\
        CORPORATE GOVERNANCE: Establishment of (x4)}
        &  Royal Bank of Canada (7\%), Banco Santander (6\%), UniCredit (6\%) 
        \\ \hline

        Global-Foundries &  \makecell{
        WATER: Use of (x6) \\
        AIR EMISSIONS: Reduction of (x7) \\
        PHILANTHROPY: Donation by (x7)}
        & Texas Instruments (7\%), PPG Industries (7\%), Samsung (6\%), Visa (6\%)
        \\ \hline

        Geely \newline Automobile & \makecell{
        PHILANTHROPY: Participation in (x5) \\
        SUPPLY CHAIN: Establishment of (x5) \\
        CORPORATE GOVERNANCE: Development \\ and implementation of (x5)}
        &  China Petroleum Chemical (7\%), Baidu (6\%), LG Display (6\%), Alibaba (5\%), Korean Air Lines (6\%)
        \\ \hline
        
        Saudi Aramco & \makecell{
        AIR EMISSIONS: Reduction of (x4) \\
        BIODIVERSITY: Protection of (x4) \\
        ENERGY: Investment in (x4)}
        &  Tokyo Gas (5\%), Royal Dutch Shell (5\%), Yamana Gold (5\%), Visa (5\%)
        \\ \hline

        Philip Morris & \makecell{
        WASTE: Continuous efforts to reduce, reuse, or recycle (x5) \\
        BIODIVERSITY: Continuing to set goals and work towards (x4) \\
        PHILANTHROPY: Investment in (x3) }
        & Croda (4\%), 3M (4\%), Coca-Cola (4\%), GlobalFoundries (4\%) Samsung (4\%), United States Steel (4\%)
        \\ \hline
    \end{tabular}
\end{table}

\subsection{Correlation analysis among company similarities}\label{subsec:correlationAnalysis}
\begin{figure}[htp]
    \centering
    \includegraphics[width = 0.8\textwidth]{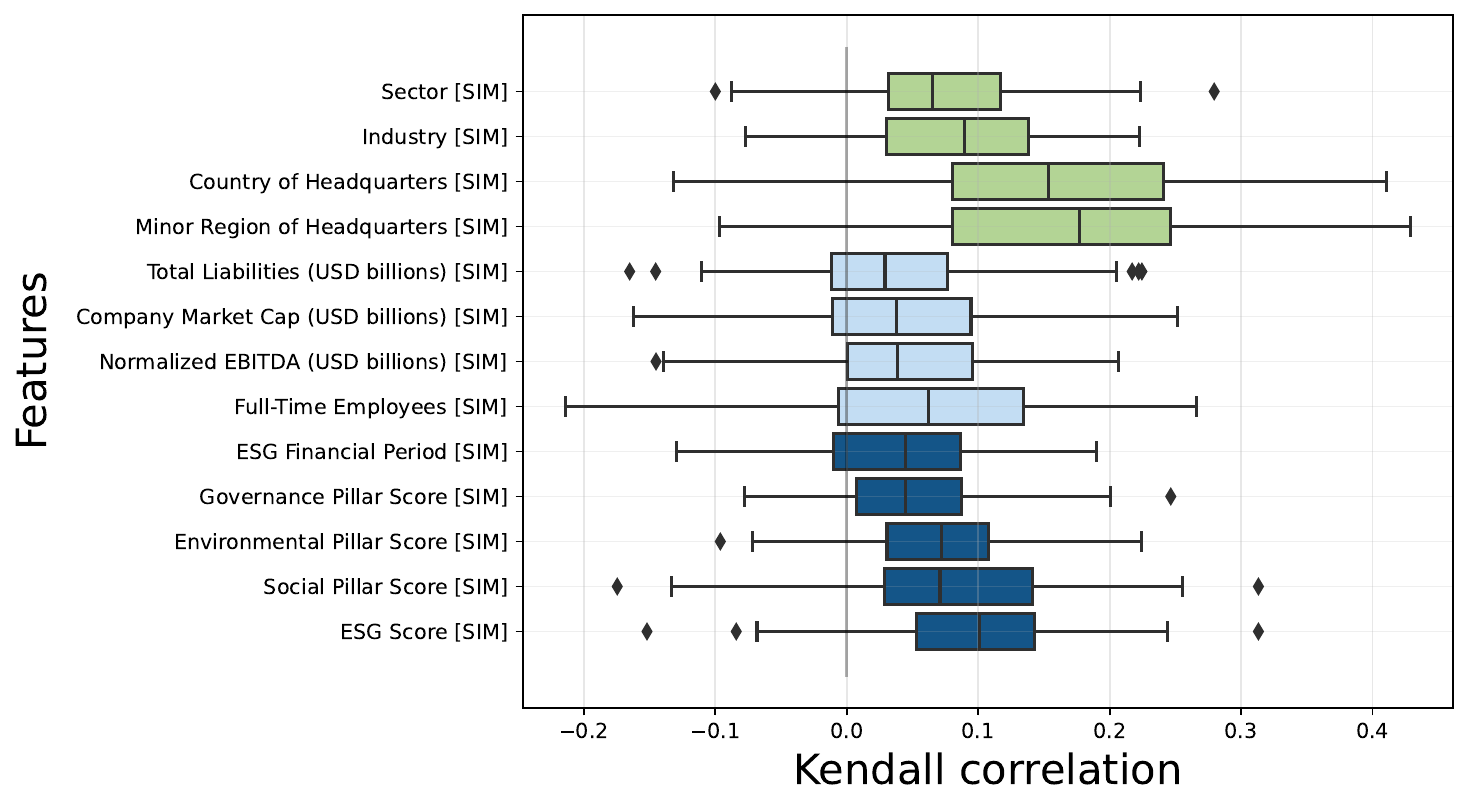}
    \caption{
    \textbf{Distributions of pairwise correlations between companies' action similarities and similarities in other company features (rows)}.
    Features are color-grouped according to their type of information.
    Light-green features categorical company characteristics, while azure and dark blue features represent numerical features concerning companies' financial and ESG information.}
    \label{fig:correlationBoxplots}
\end{figure}

This section answers the research question concerning whether company similarities in terms of jointly disclosed actions (\Cref{subsec:companySimilarity}) are associated with similarities in other company information (\Cref{subsubsec:companyInformation}).
We perform a bivariate correlation analysis using Kendall's correlation coefficient (\Cref{subsec:analysisMethods}) for each company with all information available: 81\% of the considered companies. Aggregated results are exhibited in \Cref{fig:correlationBoxplots} through box plots.

\paragraph{\textbf{Similarities in companies' disclosed actions are correlated with companies' geographical regions.}}
Action similarities have the highest, yet weak, correlation with the \texttt{Region} and \texttt{Country} of the company headquarters, with a median correlation coefficient of 0.18 and 0.15. This confirms the empirical findings discussed in the previous section (\ref{subsec:companySimilarity}).
Moreover, only these two features demonstrate median p-values, resulting from the null hypothesis test of zero monotonic correlation, below the established accepting threshold of 5\%, respectively 1\% and 2\%. 
The p-value distributions for all the features are shown in Section SM10 of the Supplementary Material (SM) document.
Taking the previous example companies, \texttt{Sony} and \texttt{Enel} exhibit a relatively high monotonic correlation between their company similarities in terms of disclosed actions and geographical locations. \texttt{Sony} has an action-country similarity correlation equal to 0.22 and an action-region correlation equal to 0.20, while \texttt{Enel} exhibits a lower action-country correlation (0.14) and a higher action-region correlation (0.25).

\paragraph{\textbf{No other statistically significant similarity correlations emerge.}}
Company similarities in \texttt{ESG scores} and \texttt{Industries} have a median pairwise correlation with companies' disclosed actions equal to 0.1 and 0.09.
Their statistical significance appears however relatively weak due to high median p-values (13\% and 15\%), which suggest accepting the null hypothesis of zero monotonic correlation.

\paragraph{\textbf{ESG scores are only correlated with their underlying components.}}
After analysing company similarities from the disclosed action perspective, we perform a pairwise monotonic correlation analysis to unveil possible confounding factors for company similarities.
Strong monotonic correlations appear between similarities in companies' \texttt{Region} and \texttt{Country} (median correlation equal to 0.7) as well as between companies' \texttt{ESG score} and their \texttt{Social} (0.5) and \texttt{Environmental Pillar scores} (0.4).
No other relevant correlations emerge for ESG scores or other company information.
A graphical representation of all the pairwise correlations is exhibited in Section SM13 of the Supplementary Material (SM) document.

\subsection{Interpretability of ESG scores} \label{subsec:ESGinterpretability}
Lastly, we investigate the interpretability of companies' ESG scores by employing a first-order linear regression and the SHAP framework (\Cref{subsubsec:ESGinterpretability}).
We specifically evaluate how various factual and corporate aspects impact these scores using features based on the companies' extracted actions (such as the most disclosed ESG topics), and additional financial and company-specific information.

\begin{figure}[htb]
    \centering
    \hspace*{2mm}
    \includegraphics[width = 1\textwidth]{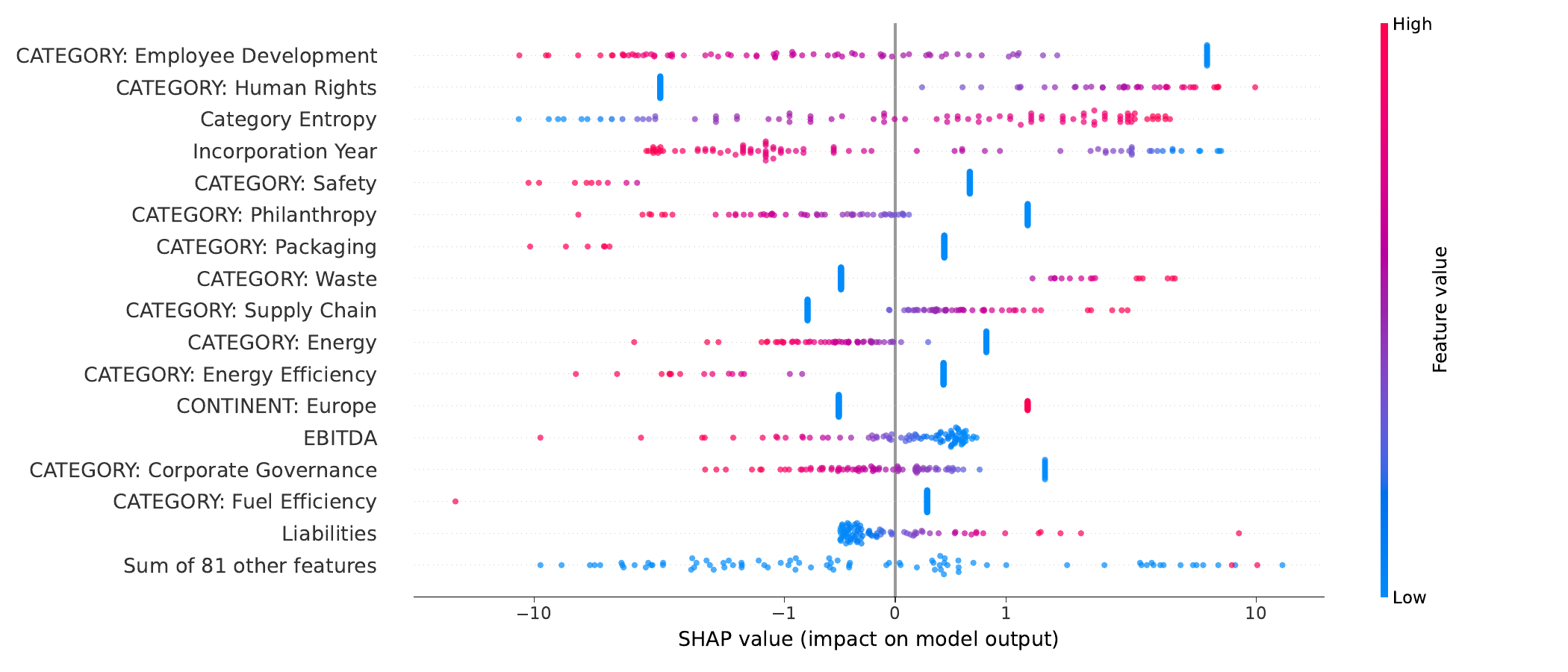}
    \caption{\textbf{Summary of the top sixteen features impacting the most the inference of ESG score.} The features are ordered according to their median shape value.
    The x-axis represents the degree of a positive and negative impact on model output. Each dot represents a company instance and colours represent the company values of the standardised feature.}
    \label{fig:explainabilityFeatureImportance}
\end{figure}

\begin{figure}[htp]
    \centering
    \includegraphics[width = 0.8\textwidth]{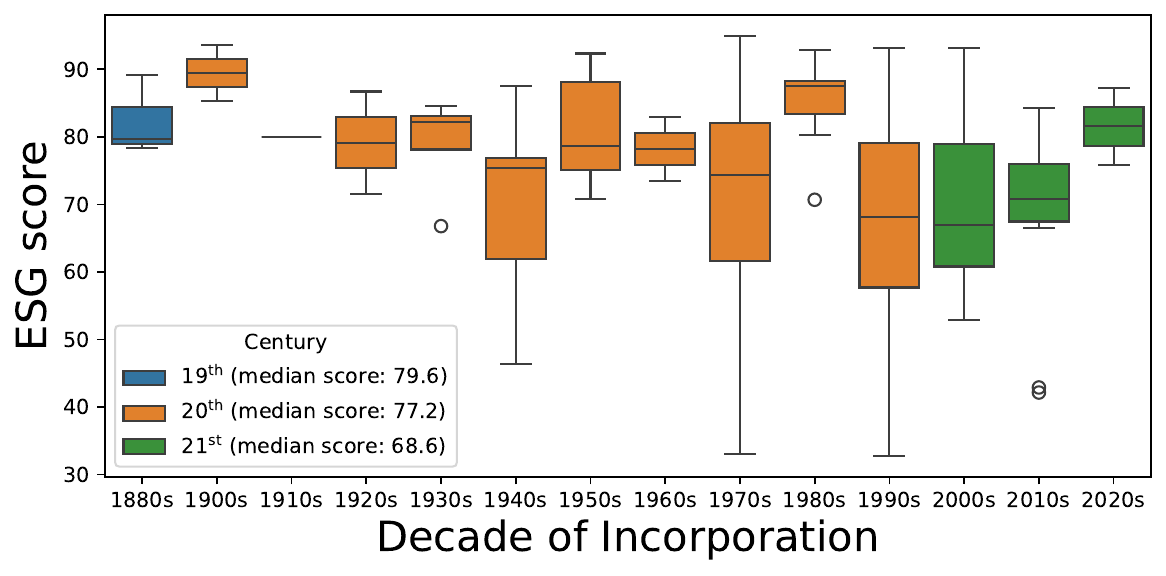}
    \caption{\textbf{Companies' ESG scores grouped by the decade of incorporation}. The ESG scores refer to the fiscal year of the companies' sustainability reports considered in our work, almost all from the 2020s.
    Colours map the century of the decades and legend entries also exhibit the median ESG score for each.}
    \label{fig:esgScoresByDecade}
\end{figure}

\paragraph{\textbf{Social-related actions, company transparency and incorporation dates have a significant impact on ESG scores.}}
On average, the most impacting aspects affecting ESG scores are the percentages of disclosed actions related to \texttt{Human Rights} and \texttt{Employee Development}, with a mean SHAP value of 2.6 and 2.7.
High percentages of the former (colour scale in \Cref{fig:explainabilityFeatureImportance}) positively impact ESG scores, while the latter has the effect of hurting them.
High disclosing percentages in actions related to \texttt{Philanthropy} (mean SHAP value of 1.1) and \texttt{Energy} (0.7) also hurt companies' scores.
Similar average magnitude, but an opposite effect, disclosing several actions related to \texttt{Waste} (0.8) or \texttt{Supply Chain} (0.7) has a positive impact.
Furthermore, high variety in the disclosed ESG topics (\texttt{Category Entropy}) positively affects ESG scores with a mean SHAP value of 2.1 (\Cref{fig:explainabilityFeatureImportance}).
Sharing a similar average magnitude (1.9), being founded earlier, represented by an older \texttt{Incorporation Year}, positively impacts a company's score.
This is also validated numerically by a negative Kendall correlation equal to -0.22 (p-value of 0.5\%) and visually by \Cref{fig:esgScoresByDecade} which groups the companies' ESG scores by their decade of incorporation.
The median ESG score of the fifty-three companies funded in the 20\textsuperscript{th} century is 77.2, whereas the thirty-one companies established in the current century exhibit a lower median score of 68.6. Notably, the three companies founded in the 19\textsuperscript{th} century exhibit the highest median ESG score equal to 79.6.

Further noteworthy factors positively impacting ESG scores are being a European company (\texttt{CONTINENT:Europe}, mean SHAP value of 0.7) and exhibiting a high level of \texttt{Liabilities} (0.5).
In contrast, high annual earnings (\texttt{EBITDA}, 0.6) have a slight negative impact on ESG scores.
The positive influence of being European can be also validated by grouping ESG scores by company region: the twenty-six companies from Europe exhibit the highest average ESG score equal to 82, the forty-nine American companies have an average ESG score of 69.7, whereas the average ESG score of the twelve Asian companies is equal to 67 (see Section SM18 in the Supplementary Material document).

\begin{figure}[htp]
    \centering
    \includegraphics[width = 0.95\textwidth]{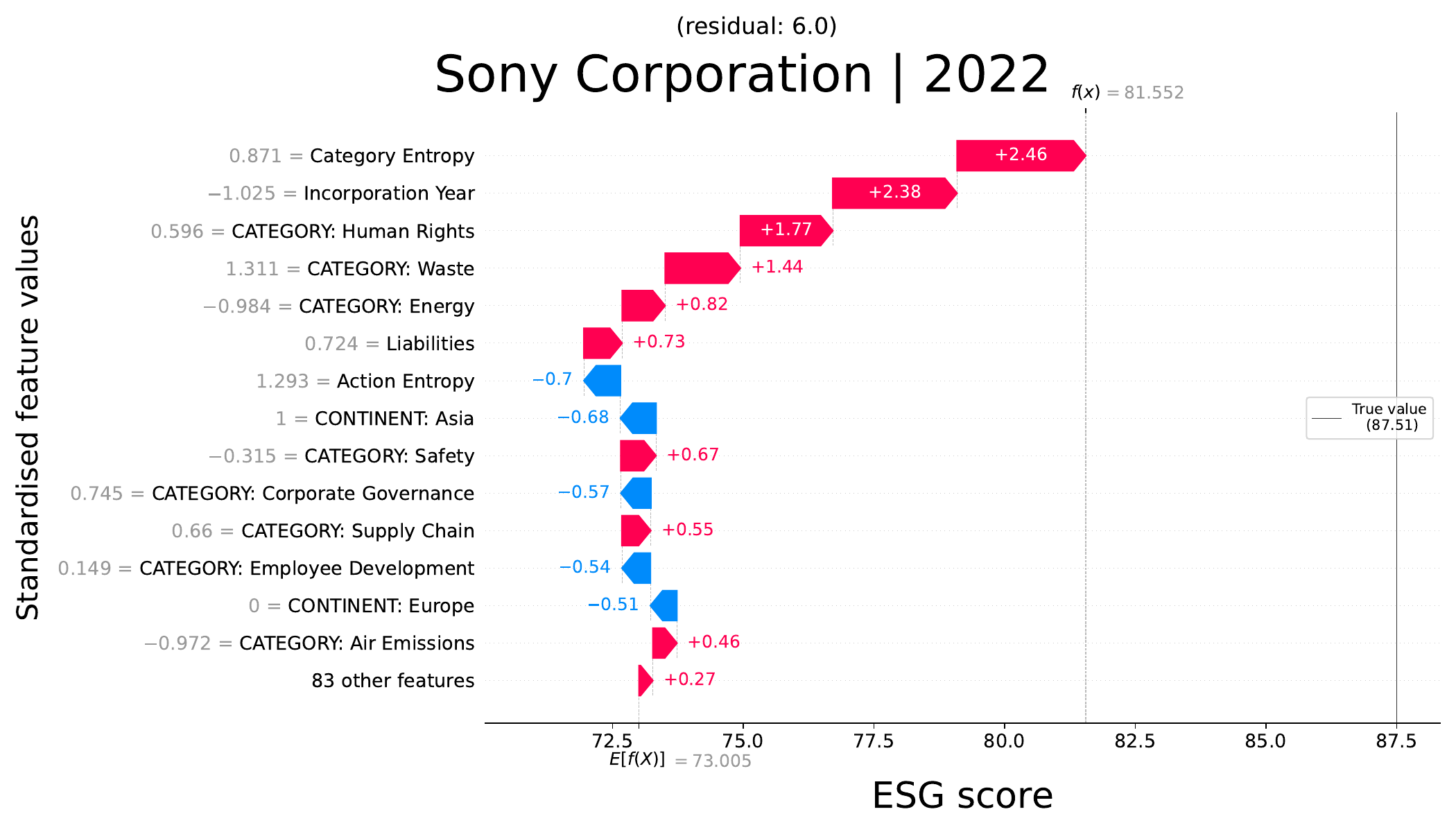}
    \caption{\textbf{Example of explanations for individual predictors for the ESG score of a company.} The category-based features are extracted from the 2022 sustainability report of the Japanese company \texttt{Sony} and other company information from the same fiscal year is considered. In addition, the actual ESG score and the model error (residual) are shown.}
    \label{fig:explainabilitySonyScore}
\end{figure}

\paragraph{\textbf{Local interpretability analysis reveals company-specific impacting factors.}}
Moving from global to local interpretability, we choose \texttt{Sony} as an example company and investigate the most impacting factors for its ESG score (\Cref{fig:explainabilitySonyScore}).
The \texttt{Incorporation Year} and \texttt{Category Entropy} features, respectively far below (standardised value of -1.03, representing the year 1946) and above (0.87, 3.6 nats) the average of company values, positively affect its score.
In addition, disclosing several actions related to \texttt{Human Rights} (0.57, 4.4\% of all its extracted actions) and \texttt{Waste} (1.31, 3.4\%) has a positive impact.
In contrast, disclosing fewer \texttt{Energy}-related actions than the average (-0.98, not among its top ten disclosed topics) positively affects its scores.
Furthermore, being an Asian company (\texttt{CONTINENT:Asia} and \texttt{CONTINENT:Europe}) slightly hurts its ESG score. 

\paragraph{\textbf{Region-based interpretability analysis reveals common patterns.}} 
We also conduct a more granular analysis by exploring the ESG score interpretability of a company cluster.
Coherently with the example company, we select Asian companies encompassing five Chinese companies (e.g., Alibaba), five Japanese companies (e.g., Sony), Aramco (Saudi Arabia) and Greely Automobile (Hong Kong).
The \texttt{Incorporation Year} strongly affects their ESG scores, with an average SHAP value of 2.7, in line with the global interpretation. 
This is further validated by a strong negative Kendall correlation of -0.61 (p-value of 0.7\%).
All companies established in the 20\textsuperscript{th} century, such as Toshiba (1904, score of 93.6), Toyota (1937, score of 84.5) and Geely Automobile (1946, score of 75.4), exhibit ESG scores above 66.
Whereas, those established in the current century have all lower scores such as Baidu (2000, score of 53.5), China Evergrande (2006, score of 52.8) and Aramco (2018, score of 42.9).
From the geographical point of view, being a Chinese company (\texttt{COUNTRY:China}) hurts ESG scores with an average SHAP value of 0.6.
This is further emphasised by the observation that all the Chinese companies consistently have ESG scores below 62.5, whereas both Hong Kong-based and Japanese companies consistently display higher scores.
Furthermore, this analysis confirms that disclosing several actions related to \texttt{Human Rights} (e.g., Toshiba and Tokyo Gas) and \texttt{Waste} (e.g., Toyota and Sony) positively impact ESG scores, whereas their absence negatively affects them (e.g., Baidu and Daikin Industries).
The ESG scores group by region, a details list of the considered Asian companies and the bee-swarm graph of the latter analysis are reported in the Supplementary Material (see Sections SM18, SM19 and SM20).

\section{Discussion} \label{sec:discussions}
Now, we address the practical implications of our findings (\Cref{subsec:praticaImplications}) as well as the methodological implications of our proposed approach (\Cref{subsec:methodologicalImplications}).
Lastly, we discuss some potential limitations of our work in \Cref{subsec:limitations}.

\subsection{Practical implications}\label{subsec:praticaImplications}

\paragraph{\textbf{High action variety in addressing ESG topics.}}
As highlighted in \Cref{subsec:bipartiteAnalysis} and \ref{subsec:diversityAnalysis}, companies address ESG topics from many perspectives, ranging from recognition and commitments to developments, partnerships and compliance.
This foregrounds the complexity and joint efforts needed to address ESG-related aspects and the involved external subjects such as regulatory agencies.
Our analysis unveils that the same action is disclosed, on average, by only 2\% of the companies, and by only 15\% of all the company sectors, confirming a lack of a common approach across companies and different sectors.
However, some ESG topics are addressed through a common strategy by the majority of the considered companies: 
the actions \texttt{Air Emission:Reduction of} and \texttt{Energy:Reduction of} are disclosed by 70\% and 61\% of the companies (\Cref{subsec:bipartiteAnalysis}).

\paragraph{\textbf{The ESG phenomenon has blurred boundaries and includes plenty of socially responsible topics.}}
Concerning the ESG topics disclosed by companies, our methodology extracts 
542 distinct ESG topics from companies' sustainability reports, representing an eight-times greater set of topics originally included in the ESG categorization exploited as a semantic reference in this work (sixty-four categories, \cref{subsubsec:esgCategorization}).
Firstly, this unveils the broad scope of the ESG phenomenon involving socially responsible topics ranging from \texttt{Waste Management} and \texttt{Supply Chain} to \texttt{Employee Safety} and \texttt{Tax Compliance}.
Secondly, this highlights the presence of widely disclosed topics, such as \texttt{Supply Chain}, and sector-focused topics such as \texttt{Packaging} and \texttt{Water} for the Consumer Staples sector.
The diversity analysis reported in \Cref{subsec:diversityAnalysis} confirms this sector-based importance for certain topics. 
For instance, the ESG topic \texttt{Water} is more stressed by water-intensive companies, while \texttt{Packaging} is more emphasised by companies producing consumer staples.
A phenomenon already emerged in the ESG literature~\cite{eccles2012need, khan2016corporate, busco2020preliminary} and referred to as ESG Industry Materiality (\Cref{subsubsec:companyInformation}), validating our data-driven insights.
In addition, the comparison with sector- and industry-level ESG materiality (\Cref{subsec:diversityAnalysis}) allows us to compare companies' disclosures with the relevant issues that were expected in their sustainability reports, spotlighting differences in topic coverage.


\paragraph{\textbf{Exogenous factors might influence companies' non-financial disclosures.}}
The findings reported in \Cref{subsec:companySimilarity} emphasise company similarities based on their sectors, confirming indirectly a relatively high presence of common strategies among companies from the same sector.
However, the most impacting factor in grouping companies based on their disclosures is their geographical region as shown in \Cref{subsec:companySimilarity} and \Cref{subsec:correlationAnalysis}. 
This represents an interesting finding from our data-driven work, which highlights the influence of the company's geographical origins on the company's disclosures, although businesses nowadays operate in a global market.
One possible explanation is that these region-based disclosing similarities may be influenced by different regulatory compliance~\cite{campbell2007would} or country factors such as the country's political, labour and cultural system~\cite{yu2021international, baldini2018role, drempetic2020influence}.
European, American and Asian companies could address their Corporate Social Responsibility (CSR) by prioritising different socially responsible efforts, investments, and disclosures based on the demands coming from their region.
For instance, Baldini \emph{et al.}~\cite{baldini2018role} found that a country's labour union density positively impacts social and governance disclosure, whereas Yu \emph{et al.}~\cite{yu2021international} unveiled that the lack of political rights has a negative influence on ESG disclosure.
On the other hand, CSR might also be influenced by regulatory agencies~\cite {campbell2007would}.
For example, European companies might prioritise environmental-related issues due to more stringent climate regulations in Europe, such as the European Union's Emissions Trading  System~\cite{EUEmissionsTrading} or the ambitious “Fit for 55" plan~\cite{Fit552023a} recently proposed by the European Union to make this region climate-neutral by 2050.
This hypothesis of a greater European commitment towards environmental aspects can also be indirectly validated by looking at the European companies' performance in the environmental pillar: European companies have the highest average environmental score of 81.8, Asian companies exhibit an average score of 69, whereas American companies exhibit the lowest average score of 65.6 (see Section SM18 in the Supplementary Material document).

\paragraph{\textbf{Companies' social and environmental performance hold greater importance than the governance performance in the combined ESG scores.}}
The bivariate correlation analysis reported in \Cref{subsec:correlationAnalysis} shows that similarities in ESG scores are neither associated with similarities in disclosed actions nor other financial or company characteristics, representing a noteworthy finding of our work.
It however unveils strong monotonic correlations between similarities in the companies' region and country as well as between the ESG score and the social and environmental pillar score (\Cref{subsec:correlationAnalysis}).
These two appear fairly trivial associations: first, the ESG score is a weighted score combining the scores of the three E/S/G pillars (\Cref{subsubsec:companyInformation}); second, the region and country have a natural geographical relation. 
However, the monotonic associations of ESG scores could be exploited to roughly infer the average influence, and thus the importance, of the E/S/G pillar scores towards the combined score. 
For instance, a weak or zero monotonic correlation suggests that (dis)similarities among scores of one specific pillar are not associated with (dis)similarities in the combined score.
This could imply a particular pillar holds relatively less importance, or weight, in determining the combined ESG scores.
Conversely, when a significant pillar is present, its (dis)similarities reflect the (dis)similarities of the combined ESG scores.
Hence, based on the monotonic associations of ESG scores, it could be inferred that, on average, the social pillar (0.5) holds slightly greater importance compared to the environmental pillar (0.4). In contrast, corporate governance bears minimal importance in ESG scores (0.2).
Although the company sample and the fiscal years considered may influence the inference of the rating agency's methodology, insights on E/S/G weightings can be helpful to validate the impacting factors of ESG scores unveiled in \Cref{subsec:ESGinterpretability}. 

%
\paragraph{\textbf{Exogenous factors can influence the quality of companies’ non-financial disclosures, indirectly impacting their ESG performance assessments.}}
The interpretability analysis of ESG scores (\Cref{subsec:ESGinterpretability}) highlights that the company's disclosures impact ESG scores more than other financial aspects or company characteristics.
Disclosing several ESG topics (\texttt{category entropy}) positively affects ESG scores, whereas fewer disclosed ESG topics hurt scores. 
This data-driven insight accordingly confirms that transparency on non-financial information rewards companies' ESG assessment~\cite{labella2019devil, drempetic2020influence}.
The analysis of \Cref{subsec:ESGinterpretability} also confirms the negligible impact of governance-related topics towards ESG scores in comparison to social- and environmental-related topics such as \texttt{Human Rights, Energy} and \texttt{Waste}.
The findings of \Cref{subsec:ESGinterpretability} also unveil that disclosing many actions related to the ESG topics \texttt{Employee Development} and \texttt{Energy} negatively impact ESG scores.
One hypothesis is that their high presence in a company's sustainability report might reveal a poor coverage of other important ESG topics~\cite{refinitivMethodology}.  
For example, in the local interpretation of Sony's ESG score, a low disclosing percentage of \texttt{Energy}-related actions and a high disclosing percentage of \texttt{Waste}-related actions positively impact its score.
Other noteworthy findings of this analysis are the impacts of the company's incorporation year and region on ESG scores.
Being founded earlier and being a European company positively impacts companies' ESG scores, whereas being a Chinese company hurts them (\Cref{subsec:ESGinterpretability}).
The latter factor might be reinforced by the fact that all the considered Chinese companies were funded between 1999 and 2006, likely due to the remarkable economic development of this region starting in the 2000s, and thus associated with the negative impact of being relatively young companies (incorporation year).
However, the impact of the company's region on ESG scores is coherent with the region-based disclosing similarity previously highlighted and validated by ongoing discussions in the ESG literature.
The study of LaBella \emph{et al.}~\cite{labella2019devil} validates this European bias as well as the important role of regulatory agencies.
The authors~\cite{labella2019devil} discovered a bias among rating agencies when evaluating ESG performance, showing a preference for European companies over their North American, Emerging Markets, and Developed Asian counterparts.
They proposed that this bias could stem from variations in formal reporting requirements across different jurisdictions, contributing to differences in the quality of companies' non-financial disclosures~\cite{labella2019devil}.
This indirectly spotlights the pioneering and benchmark role of European companies concerning non-financial disclosures and CSR. 
Additionally, there are studies~\cite{drempetic2020influence, labella2019devil, dobrick2023size} that suggest non-financial disclosures, and indirectly, companies' ESG assessments, might also suffer from biases based on company size. 
This is because generating non-financial disclosures can be both financially and labour-intensive~\cite{labella2019devil}. 
As a result, larger companies could invest more economic and human resources in improving their transparency on non-financial aspects, leading to a positive influence on their ESG evaluation.
Our interpretability analysis (\Cref{subsec:ESGinterpretability}) unveils negligible evidence of this company size bias: a greater number of full-time employees has a positive, yet marginal, impact on ESG scores (average SHAP value of 0.24), whereas the company's market capitalization has no impact on interpreting companies' ESG scores.
In addition, the Kendall pairwise correlations of company similarities (\Cref{subsec:correlationAnalysis}) concerning these two company variables and similarities in ESG scores are not statistically significant: a monotonic correlation of 0.1 for the number of employees (p-value of 71\%), and a zero correlation for the market capitalization (p-value of 78\%).

\subsection{Methodological implications}\label{subsec:methodologicalImplications}

As mentioned in \Cref{subsec:RetrievalAugmentedGeneration}, generative LLMs provide us with the semantic understanding and flexibility needed to overcome the limitation of traditional OIE approaches which rely only on the syntactical sentence structure.
This allows us to generate semantically aware and ESG-focused triples instead of traditional SPO ones. 
This is pivotal in generating all the meaningful ESG-related insights of our work. 

\paragraph{\textbf{
Utilising generative LLMs and the ESG categorization for semantic guidance enhances retrieving more comprehensive data-driven insights.}}
The flexibility and generative abilities of these generative language models also allow us to highlight, and overcome some limitations in the data sources such as those of the ESG categorization used in our work (\Cref{subsubsec:esgCategorization}).
This categorization extrapolates a concise set of ESG topics by categorising several ESG-related indicators shared among ESG rating providers.
It accordingly derives the scope of the ESG phenomenon from the perspective of rating agencies, a different viewpoint compared to companies' disclosures analysed in our work.
However, these different points of view, and the methodology based on generative LLMs, help us to unveil differences among the ESG topics considered by rating agencies and disclosed by companies.
We indeed extract a set of ESG topics/categories disclosed in companies' sustainability reports that is eight times larger than the original list of categories of this classification (542 versus 64, \Cref{subsec:praticaImplications}).
For instance, our methodology unveils \texttt{"Education"} as a pivotal ESG topic disclosed by more than two-thirds of the selected companies. 
This topic is not explicitly included in the categorization, although it could be framed within three of its categories: Access to Basic Services, Human rights (Art. 26) or Philanthropy. 
Additional examples encompass the extracted ESG category “\texttt{Circular Economy}" which might fall under the categorization categories of “Waste" or “Resource Efficiency" as well as the ESG category “\texttt{Air Quality}" which could be framed within “Green Buildings" or “Health and Safety".

Accordingly, the aforementioned ESG categorization encompasses critical topics hidden within vague categories or potentially overlooks them altogether, resulting in a reduction of substantial significance in subsequent analyses.
Our approach addresses this limitation by leveraging a generative LLM in conjunction with the ICL technique and the RAG paradigm.
This allows us to jointly simulate the outputs of a supervised text classifier, whose labels are the categories of the ESG categorization, and semantically generalize those labels.
The ESG categorization is consequently leveraged as semantic guidance by the LLM, helping itself to extract more suitable topics while keeping the domain and semantics of the original ESG categorization.
This semantic generalization can also diminish the reliance on specific ESG categorizations when classifying sentences since the ESG categories are exploited as semantic references rather than fixed labels.
Nevertheless, this could also lead to the undesirable phenomenon of over-specialization which was tackled using semantic clustering (\Cref{subsec:knowledgeGeneration}). 
We also used this ESG categorization, in the data preparation phase, to filter the report sentences using the text embeddings (\Cref{subsubsec:nlppipeline}). This semantic-oriented filtering approach allows us to further move towards a taxonomy-agnostic methodology as filtering is based on semantics rather than single keywords. 


\paragraph{\textbf{Extracting insights from companies' sustainability reports using generative LLMs, and graph representations.}}
Lastly, our methodology differs from other recent ESG-focused and LLM-based tools (\Cref{sec:relatedWork}) such as ChatClimate~\cite{vaghefi2023chatclimate} and ChatReport~\cite{ni2023paradigm} by employing the paradigm of Retrieval-Augmented Generation (RAG), alongside In-Context Learning, for Knowledge Graph generation. 
This methodology, in combination with bipartite graph representation, allows us to report meaningful insights concerning the actions disclosed in companies' sustainability reports. 
In comparison, ChatClimate adopts the RAG paradigm to augment ESG-related questions for question-answering, whereas ChatReport leverages this paradigm to operationalise the compliance assessment of sustainability reports towards the recommendation guidelines of the Task Force on Climate-related Financial Disclosures (TFCFD).

\subsection{Limitations} \label{subsec:limitations}
Our data preparation NLP pipeline relies on a PDF parser~\cite{pypiPyMuPDF} to extract texts from sustainability reports.
This parser extracts all texts including those from infographics and tables.
This might yield some sentences without a proper syntactic structure, making extracting semantic meaning from them difficult or even impossible.
\Cref{tab:topKSentences} in \Cref{subsub:semanticSearch} exhibits an example of this issue in the first sentence retrieved for the ESG category \texttt{Waste}.
Although this sentence contains some details regarding this topic, it lacks a coherent message.

However, the semantic understanding of LLMs in combination with the In-Context Learning technique and the paradigm of RAG could implicitly address this issue. 
Indeed, the sentence coverage of our retrieval-augmented triple generation (\Cref{subsec:RetrievalAugmentedGeneration}) is equal to 68.1 \%, meaning that the language model acts as an implicit filtering layer and avoids generating triples for just about 30\% of all the processed input sentences.
The aforementioned example is within this set of ignored sentences.
Although an end-to-end approach might be desired, discarding such meaningless sentences beforehand could help avoid an unnecessary computational workload. 
For instance, future works could tackle this issue by enhancing document parsing (e.g., by preserving the original layout) or adding a further, yet lightweight, filtering component.
The latter might filter sentences according to their syntactical correctness or meaningfulness. 

Another potential limitation concerns the interpretability of ESG scores using SHAP values. 
The SHAP framework is used to roughly interpret the impact of predictors on individual predictions. 
Global interpretability is derived using simple aggregating statistics such as mean/median SHAP values.
Nevertheless, this aggregating approach for global interpretability might result in a mixed global interpretation in the presence of high diversity in the observations as can be a set of companies from worldwide nations covering eleven distinct sectors.
The global impact of some predictors could still be accurate, yet some might be a mix of sector-dependent relationships or caused by the diversity of cause-effect connections.
Indeed, the region-based interpretability analysis (\Cref{subsec:ESGinterpretability}) unveils more impacting factors or relationships for a specific company cluster in comparison to the global interpretability (\Cref{subsec:praticaImplications}).
However, future works might conduct a further subset-based interpretability analysis by adopting a bottom-up approach and letting company groups emerge by themselves. 

Lastly, the data provider for ESG scores used for our work might be a limitation worth highlighting.
We rely on the ESG scores from the Refinitiv platform (\Cref{subsubsec:companyInformation}), but, as highlighted in the Introduction Section, rating agencies have their assessment methodologies which could result in divergences in companies’ ESG scores.
Consequently, the findings relying on ESG scores (\Cref{subsec:correlationAnalysis} and \ref{subsec:ESGinterpretability}) might vary using ESG scores from other rating agencies such as Sustainalytics which adopts a risk-based assessment~\cite{sustainalyticsMethodology}.
In addition, future works could integrate further ESG-related attributes from these rating agencies such as quantifying companies' water withdrawal, hazardous waste, gender pay gap and employee turnover.


\section{Conclusions} \label{sec:conclusions}
In this work, we proposed a data-driven methodology based on generative LLMs to systematically evaluate the context in which ESG topics are disclosed by companies in their sustainability reports.
The objective of this work was to contribute to the emerging field of 
automatic information extraction from companies' sustainability reports by implementing the best NLP pipeline to extract structured insights from lengthy and visually rich PDF documents.
This generative LLM-based approach allowed us to directly investigate the companies' perspective concerning the ESG phenomenon.

Large Language Models (LLMs) can be versatile tools to accomplish diverse NLP-related tasks including also extracting structured information from textual data.
We further explored this promising research direction by adopting the Retrieved-Augmented Generation (RAG) paradigm, alongside the In-Context Learning (ICL) technique, to extract ESG-related information as semantically structured triples.
We then adopted a graph representation (bipartite graphs) to extract non-trivial statistics and conduct meaningful analyses concerning companies' disclosed actions.
We employed a pre-trained language model from the open-source community, distinguishing us from other recent publications as far as we know.
Furthermore, our LLM-based methodology overcomes important limitations related to traditional OIE techniques and ESG categorization, allowing us to generate both semantically aware and ESG-oriented triples instead of traditional subject-predicate-object (SPO) triples.
This helped us to report meaningful findings such as statistical, similarity and correlation analyses on the ESG topics and actions extrapolated from companies' sustainability reports as well as conduct an interpretability analysis of ESG scores.
Future works might integrate further data sources, such as ESG-related news, to analyse possible inconsistencies in companies' claims and actions.
Another interesting research direction might be to integrate Semantic Role Labelling (SRL) to enhance the extracted structured information with semantic roles, such as the agent, manner, and purpose of an action, as well as other contextual information, such as time and location.

\section*{List of Abbreviations}
\textbf{CSR:} Corporate Social Responsibility \\
\textbf{EBITDA:} Earnings Before Interest, Taxes, Depreciation, and Amortization \\
\textbf{ESG:} Environmental, Social, and Corporate Governance \\
\textbf{GICS:} Global Industry Classification Standard \\
\textbf{ICL}: In-Context Learning \\
\textbf{KG(s):} Knowledge Graph(s) \\
\textbf{LLM(s):} Large Language Model(s) \\
\textbf{NLP:} Natural Language Processing \\
\textbf{OIE:} Open Information Extraction \\
\textbf{OLS:} Ordinary Least Squares  \\
\textbf{RAG:}  Retrieved Augmented Generation \\
\textbf{SASB}: Sustainability Accounting Standard Boards \\
\textbf{SDG(s):} Sustainable Development Goal(s) \\
\textbf{SHAP:} SHapley Additive exPlanations  \\
\textbf{SM:}  Supplementary Material \\
\textbf{SPO:} Subject-Predicate-Object \\

\section*{Declarations}

\subsection*{Supplementary Material}
Further tabular and graphical views are exhibited in the 
\href{https://drive.google.com/file/d/177oOFNoJwECCbGyDdBL8YwjHZP-YJ7Jd}{Supplementary Material (SM) document}.

\subsection*{Availability of data and materials} 
The sustainability reports used, processed and analysed during the current study can be publicly retrieved from the following websites: 
\href{https://sasb.org/company-use/sasb-reporters}{sasb.org}~\cite{sasbReports} and 
\href{https://responsibilityreports.com}{responsibilityreports.com}~\cite{responsibilityReports}.
The complete list of the companies we considered can be found in Section SM23 of the supplementary material document.
The ESG categorization exploited in our work was extrapolated from Table IV (4) of the work~\cite{berg2022aggregate} by Berg \emph{et al.}. It is also exhibited in Section SM21 of the supplementary material document.
The sector-level materiality was extracted from the table in Appendix C of the work~\cite{khan2016corporate} by Khan \emph{et al.}
The ESG scores and other financial information used in this study are available from the Refinitiv platform, but restrictions apply to the availability of these data, which were used under license for the current study, and so are not publicly available. 
All the datasets used, processed and analysed and the extracted companies triples can be found in this repository: \href{https://github.com/saturnMars/derivingStructuredInsightsFromSustainabilityReportsViaLargeLanguageModels}{github.com/saturnMars/derivingStructuredInsightsFromSustainabilityReportsViaLargeLanguageModels}.
Further data and findings are available from the corresponding author upon reasonable request.


\subsection*{Funding}
The work of JS has been partially funded by Ipazia S.p.A.  
BL and AP acknowledge the support of the PNRR project FAIR - Future AI
Research (PE00000013), under the NRRP MUR program funded by the NextGenerationEU.

\subsection*{Author's contributions}
MB collected, prepared and processed the data, implemented the proposed approach, wrote the paper and interpreted the findings. 
MB, CN and JS conceptualised and designed the proposed approach.
CN, JS, BL and AP supervised the research direction of this study.
CN, JS and BL supervised the writing process.
All authors read and approved the final manuscript.


\bibliographystyle{unsrt}  
\bibliography{main}  

\appendix 

\section{Qualitative analysis of the generated triples} \label{apx:qualitativeAnalysis}
We evaluated the generated triples by prompting the same LLM (WizardLM~\cite{responsibilityReports}) to evaluate triple quality.
We prompted the model to evaluate the coherence and alignment between the structured information (output) and the sentence (input), considering also the coherence of each triple attribute (\texttt{cat, pred,} and \texttt{obj}).
The model was prompted to evaluate each, leveraging also its ICL abilities, using numerical scores on a scale from 0 to 3.
The full model instruction used for this evaluation is exhibited in Section 25 in the Supplementary Material (SM) document.
We specifically analysed a random sample of 1,000 triples from the total of forty thousand triples generated from all companies' sustainability reports. 
The triple sample has an average score of 2.65 (std: 0.44), showing a fairly high quality of the structured information extracted.
Specifically, the ESG category (\texttt{cat}) and object (\texttt{obj}) show high average performance (2.76 and 2.78) in comparison to the generated predicate (\texttt{pred}, 2.5).
The distributions are visually exhibited in Section 26 in the Supplementary Material (SM) document.
\Cref{tab:tripleComparsions} showcases five generated triples with their evaluation. 
The ESG topic mentioned directly or indirectly in the sentence is semantically captured fairly well, whereas extracting the predicate affecting it might be more challenging. 
For example, the predicate of the last triple in \Cref{tab:tripleComparsions} displays low quality in terms of syntax and coherence with the action of the original sentence. 
For instance, the predicate “\texttt{Creation of}" could have been more suitable for this sentence.
However, this triple notably captures the actual main ESG topic of this tricky sentence (\texttt{Supply Chain}), despite the mention of another potential, yet secondary ESG aspect (employment opportunities).
Furthermore, the second triple in \Cref{tab:tripleComparsions} displays another phenomenon observed in some triples: lengthy predicate attributes enveloping direct objects.
This happens especially when the disclosed actions involve transitive verbs as in this example (to meet something).
Thus, despite its semantic correctness and coherence with the disclosed action, this might make our effort in generalising challenging.

\begin{table}[htb]
    \centering
    \caption{\scriptsize\textbf{Evaluation of a sample of five generated triples and their original sentences.}
    The triples were automatically evaluated by an LLM outputting a 0-3 numerical score for each triple attribute.}
    \resizebox{\columnwidth}{!}{
    \scriptsize
    \begin{tabular}{m{8cm}|c|c|c|c|c} 
        \textbf{Sentence} & \textbf{Generated triple} & \textbf{\texttt{cat}} & \textbf{\texttt{pred}} & \textbf{\texttt{obj}} & \textbf{AVG} \\ \hline

         We evaluate products and materials against HSE compliance, safety and sustainability metrics &
        \makecell{ 
         \{'esg\_category': 'Product Safety',
         'predicate': 'Evaluation of'\\
         'object': 'Products and materials against HSE compliance, safety and sustainability metrics'\}} & 3 & 3 & 3 & 3 \\ \hline

        With increasing digitalisation, the ability to provide a data environment meeting the highest security and regulatory standards, such as GDPR, is critical &
        \makecell{ 
         \{'esg\_category': 'Data Privacy' \\
         'predicate': 'Meeting the highest security and regulatory standards'\\
         'object': 'The ability to provide a data environment'\}} & 3 & 3 & 3 & 3 \\ \hline

        We partner with trusted, leading non-profit organizations committed to impacting the areas closest to our homes and our hearts, with a particular focus on Los Angeles and New York City, where we have our largest footprint &
        \makecell{ 
         \{'esg\_category': 'Philanthropy',
         'predicate': 'Partnership with'\\
         'object': 'Trusted, leading non-profit organizations'\}} & 3 & 2 & 3 & 2.67 \\ \hline
        
        That data will help us refine operations, modify purchasing decisions and meet our emission reduction goals &
         \{'esg\_category': 'Environmental', 
         'predicate': 'Use of',
         'object': 'Data'\} & 2 & 2 & 2 & 2 \\ \hline

        We also indirectly create employment opportunities for many young people in our supply chain and retail distribution networks &
        \makecell{ 
         \{'esg\_category': 'Supply Chain',
         'predicate': 'Indirect'\\
         'object': 'Employment opportunities for many young people'\}} & 2 & 1 & 2 & 1.67 \\ \hline
        
    \end{tabular}}
    \label{tab:tripleComparsions}
\end{table}

\section{Ablation study on the model instruction} \label{apx:ablationStudy}
Furthermore, since the model instruction has a great impact on the quality of the generated text~\cite{zhao2021calibrate}, we conduct an ablation study to compare qualitatively the triples generated through different prompt templates: 
with(out) In-Context Learning (\texttt{EXAMPLES} in \Cref{fig:promptTempate}) and with(out) the semantic output schema (\texttt{DATASCHEMA}).

We find that including \texttt{examples} in the prompt, and thus exploiting the In-Context Learning capabilities of the model, generates better triples in terms of both information completeness and semantic representation, observable respectively in the first two comparisons and the third one in \Cref{tab:tripleComparsion_wExamples}.
Furthermore, that helps the model compose its response by adhering to a specific output format. 
Indeed, although the model was already prompted to generate text as a valid JSON object, it outputs texts in a valid format only after adding In-Context Learning.
\begin{table}[htb]
    \centering
    \caption{\scriptsize Three comparing examples of the triples generated with/without using In-Context Learning in the model instruction.}
    \resizebox{\columnwidth}{!}{ 
    \scriptsize
    \begin{tabular}{
    >{\centering\arraybackslash}m{2.25cm}|
    >{\centering\arraybackslash}m{3cm}|
    >{\centering\arraybackslash}m{2cm}|
    >{\centering\arraybackslash}m{4cm}}
        & \textbf{ESG category} (\texttt{cat}) & \textbf{Predicate} (\texttt{pred}) & \textbf{Object} (\texttt{obj}) \\ \hline
        \texttt{EXAMPLES} & Work-Life Balance & Provision of & A policy providing clarity around flexible work options \\ \hline
        without \texttt{EXAMPLES} & Family Friendly Policies & Provides & Flexible work option \\ \specialrule{1.2pt}{1pt}{1pt}
        
        \texttt{EXAMPLES} & Health and Safety & Creation of & COVID-19 safety tips for ridesharing \\ \hline
        without \texttt{EXAMPLES} & Public Health & Created & COVID-19 safety tips \\ \specialrule{1.2pt}{1pt}{1pt}

        \texttt{EXAMPLES} & Climate Risk Management & Partnership with & CoGo to provide personalised carbon footprints  \\ \hline
        without \texttt{EXAMPLES} & Climate Risk Management & Announced & Partnership \\ \hline
    \end{tabular}}
    \label{tab:tripleComparsion_wExamples}
\end{table}

Moreover, adding a semantic output schema in the prompt (\texttt{DATASCHEMA}) 
helps the generative language model to better focus on ESG-related information as exhibited in \Cref{tab:tripleComparsion_wDataSchema}.
The semantic schema provides the language model with detailed semantic descriptions concerning the types of information to extract: “an issue related to an ESG aspect" (\texttt{cat}), “a nominalised verb affecting that aspect" \texttt{pred}, and “an entity undergoing the predicate" (\texttt{obj}).
This could drastically affect the structured information extracted from a sentence, especially in those with multiple causes, such as in the second comparison shown in \Cref{tab:tripleComparsion_wDataSchema}.
Furthermore, defining a semantic schema allows us to incorporate a list of ESG topics (ESG categorization, \Cref{subsubsec:esgCategorization}) into the semantic description of the ESG category attribute (\texttt{cat}).
This enhanced the model's ability to extract an ESG topic (\texttt{cat}) mentioned indirectly in a sentence by jointly leveraging this list of ESG categories as semantic references, its semantic understanding, and its generative abilities.
An example of this enhancement can be observed in the third comparison in \Cref{tab:tripleComparsion_wDataSchema}, while the complete semantic schema can be seen in the Supplementary Material document within the full model instruction (see Section SM2). 

\begin{table}[htb]
    \centering
    \caption{\scriptsize Three comparing examples of triples generated with/without the semantic output data schema in the model instruction.}
    \resizebox{\columnwidth}{!}{
    \scriptsize
    \begin{tabular}{
    >{\centering\arraybackslash}m{1.4cm}|
    >{\centering\arraybackslash}m{2.3cm}|
    >{\centering\arraybackslash}m{2.05cm}|
    >{\centering\arraybackslash}m{5.43cm}}
        & \textbf{ESG category} (\texttt{cat}) & \textbf{Predicate} (\texttt{pred}) & \textbf{Object} (\texttt{obj}) \\ \hline
        \texttt{DATASCHEMA} & Work-Life Balance & Provision of & A policy providing clarity around flexible work options \\ \hline
        without \texttt{DATASCHEMA} &  Work-Life Balance & Providing clarity around & Flexible work options available to parents and caregivers \\ \specialrule{1.2pt}{1pt}{1pt}
        
        \texttt{DATASCHEMA} & Philanthropy & Partnership with & The Bill and Melinda Gates Foundation and the Western Cape Department of Health \\ \hline
        without \texttt{DATASCHEMA} & Health Care & Delivery of & Over 1.4 million prescriptions \\ \specialrule{1.2pt}{1pt}{1pt}

        \texttt{DATASCHEMA} & Climate Risk Management & Partnership with & CoGo to provide personalised carbon footprints \\ \hline
        without \texttt{DATASCHEMA} & Carbon Footprint & Partnership with & To provide personalised carbon footprints \\ \hline
    \end{tabular}}
    \label{tab:tripleComparsion_wDataSchema}
\end{table}

\section{Hyper-parameter choice of the Large Language Model} \label{apx:LLMParameters}
Large Language models have some hyper-parameters for tuning their textual outputs, and consequently, some choices should be made to address these further degrees of freedom.
The temperature parameter controls the randomness of the model responses by influencing the model's confidence in its most likely output~\cite{tunstall2022natural}.
During the decoding phase, this parameter alters the model output by scaling the logits before applying the softmax function. 
A high temperature makes the model output more diverse and creative but also more unpredictable. 
Conversely, a lower temperature makes the model output more deterministic and focused.
Setting a temperature equal to zero corresponds to greedy decoding~\cite{tunstall2022natural}.
Accordingly, we opt for greedy decoding to ensure deterministic outputs and to make the generation process adhere to instructions as much as possible~\cite{ni2023paradigm}.

Another hyper-parameter that affects text generation is the beam number representing the number of tokens considered during the beam search algorithm~\cite{tunstall2022natural}. 
Beam search is a sampling decoding algorithm that improves the output of LLMs by pruning off bad thinking patterns at generation time.
This algorithm works by iteratively generating a sequence of $b_{dim}$ tokens, and then outputting the sequence with the highest probability~\cite{hewitt2022truncation}. 
We found through extensive experiments that the beam number ($b_{dim}$) ranging from 4 to 6 strikes a good balance between semantic representation and computational workload. 
We accordingly adopt a beam number equal to 6.



\section{Empirical experiments with different Large Language Models} \label{apx:experimentsLLMs}
We conducted empirical experiments using various instruction-tuned Large Language Models (LLMs): Google’s \texttt{Flan-T5-Large, Alpaca-LoRA-7B, WizardLM-7B} and OpenAI's \texttt{ChatGPT} in its version based on GPT-3.5.
These experiments aimed to assess the performance of different LLMs and identify a suitable, cost-free generative LLM for constructing an NLP pipeline to extract structured insights from ESG-related textual documents. 
Additionally, we evaluated OpenAI’s commercial ChatGPT to compare open-sourced LLMs against the state-of-the-art, general-purpose and paid Large Language Model. 
To maintain consistency, we employed the same model instruction for all LLMs which was also the one adopted for this work (Section SM2 of the Supplementary Material (SM) document).

\Cref{tab:outputComparisons} presents three comparative analyses of textual outputs generated by different LLMs, using a sample of sentences extracted from companies’ sustainability reports.
OpenAI's \texttt{ChatGPT} demonstrated exceptional performance, nearly surpassing all open-sourced LLMs in generating ESG-oriented triples and serving as the de facto golden standard.
However, in the third comparison (\Cref{tab:outputComparisons}), \texttt{ChatGPT}'s output was subjectively inferior to that of WizardLM.
Although its extracted triple conveys the meaning of the company's disclosure, we argue that it did not focus on the main action disclosed in the sentence.
It initially identified “green steel" as the main ESG-related issue ("Green Products") rather than the investment in a platform for customers; misclassifying the ESG topic of the disclosure. 
\texttt{WizardLM}, the LLaMA-based LLM chosen for this work, exhibited remarkable performance in understanding the NLP task and extracting ESG-focused structured information, closely matching the outputs from the paid OpenAI's model (\Cref{tab:outputComparisons}).
\texttt{Alpaca}, another LLaMA-based LLM, demonstrated a good understanding of the NLP task but lagged in the quality of structured information extraction across all comparisons, especially for predicates and objects (\Cref{tab:outputComparisons}).
Lastly, \texttt{Flan-T5}, Google's instruction fine-tuned version of the Text-to-Text Transfer Transformer (T5) Language Model, generated nonsensical texts, indicating a lack of understanding of the NLP task.


\begin{table}[b]
    \centering
     \scriptsize
        \caption{\scriptsize{ \textbf{Three comparisons of the outputs generated by different Large Language Models}. The original sentence is exhibited before the different model outputs. All the LLMs are prompted with the same model instruction adopted in this work. Model outputs are reported in their raw form.}}
    \resizebox{\columnwidth}{!}{
    \begin{tabular}{>{\centering\arraybackslash}m{1.9cm}|>{\centering\arraybackslash}m{15cm}}
        \textbf{LLM} & \textbf{Original sentence/Model output}  \\ \hline
        \multicolumn{2}{>{\centering\arraybackslash}m{\textwidth}}{
        \textit{“In 2022, we will be completing a review of our supply chain responsibility and risk assessment to identify opportunities to engage with our suppliers and build capacity."}} \\ \hline
         
         Flan-T5-Large & \texttt{json$>$src:src.getres} \\ \hline

         Alpaca-LoRA-7B & \texttt{$<$json$>$\{"esg\_actions": [\{"esg\_category": \textbf{"Supply Chain"}, "predicate": \textbf{"Review of our supply chain responsibility and risk assessment"}, "object": \textbf{"Opportunities to engage with our suppliers and build capacity"}\}]\}$<$/json$>$} \\ \hline
         
         WizardLM-7B & \texttt{$<$json$>$\{"esg actions": [\{"esg\_category": \textbf{"Supply Chain"}, "predicate": \textbf{"Review of"}, "object": \textbf{"Our supply chain responsibility and risk assessment"}\}]\}$<$/json$>$} \\ \hline

         ChatGPT (GPT-3.5) & \texttt{$<$json$>$\{"esg actions": [\{"esg category": \textbf{"Supply Chain"}, "predicate": \textbf{"Completion of"}, "object": \textbf{"A review of our supply chain responsibility and risk assessment"}\}]\}$<$/json$>$} 
         \\ \specialrule{1.2pt}{1pt}{1pt}

         \multicolumn{2}{>{\centering\arraybackslash}m{\textwidth}}{
        \textit{“Many of our buildings also capture rainwater that is used when irrigation is required."}} \\ \hline
         
         Flan-T5-Large & \texttt{json$>$separateby:entity, se} \\ \hline

         Alpaca-LoRA-7B & \texttt{$<$json$>$\{"esg\_actions": [\{"esg\_category": \textbf{"Water"}, "predicate": \textbf{"Capture of rainwater"}, "object": \textbf{"Used when irrigation is required"}\}]\}$<$/json$>$} \\ \hline
         
         WizardLM-7B & \texttt{$<$json$>$\{"esg actions": [\{"esg\_category": \textbf{"Water"}, "predicate": \textbf{"Capture of"}, "object": \textbf{"Rainwater that is used for irrigation"}\}]\}$<$/json$>$} \\ \hline

         ChatGPT (GPT-3.5) & \texttt{$<$json$>$\{"esg actions": [\{"esg category": \textbf{"Water"}, "predicate": \textbf{"Capture of"}, "object": \textbf{"Rainwater for irrigation"}\}]\}$<$/json$>$} 
         \\ \specialrule{1.2pt}{1pt}{1pt}

         \multicolumn{2}{>{\centering\arraybackslash}m{\textwidth}}{
        \textit{“This investment is a platform to provide our customers with more of the green steel they expect from like-minded partners like U. S. Steel."}} \\ \hline 
         
         Flan-T5-Large & \texttt{json$>$str$>$Investment:str$>$} \\ \hline

         Alpaca-LoRA-7B & \texttt{$<$json$>$\{"esg\_actions": [\{"esg\_category": \textbf{"Energy"}, "predicate": \textbf{"This investment"}, "object": \textbf{"Green steel"}\}]\}$<$/json$>$} \\ \hline
         
         WizardLM-7B & \texttt{$<$json$>$\{"esg actions": [\{"esg\_category": \textbf{"Environmental Management System"}, "predicate": \textbf{"Investment in"}, "object": \textbf{"A platform to provide customers with more green steel"}\}]\}$<$/json$>$} \\ \hline

         ChatGPT (GPT-3.5) & \texttt{$<$json$>$\{"esg actions": [\{"esg category": \textbf{"Green Products"}, "predicate": \textbf{"Platform to provide"}, "object": \textbf{"Customers with more of the green steel"}\}]\}$<$/json$>$}
         \\ \specialrule{1.2pt}{1pt}{1pt}

    \end{tabular}}
    \label{tab:outputComparisons}
\end{table}

\section{Performance of the OLS model} \label{apx:OLSPerfomance}
We here report the performance of the OLS regression through different metrics (see also Section SM15 in the Supplementary Material document).
Firstly, the Coefficient of Determination ($R^2$,~\cite{zhang2017coefficient}) measures the proportion of variation in the dependent variable explained by the model predictors, representing the goodness of the inference ability of the model.
Low coefficients express a little variation proportion explained by the model predictors, resulting in poor performance on the inference of the dependent variable (ESG scores). 
In contrast, in the presence of a high variation proportion explained, the model predicts the dependent variable with small errors. 
Our OLS model achieves a $R^2$ of $0.71$ using the optimal alpha, demonstrating a broad variation explained by our features to infer ESG scores.
On the other hand, the Root Mean Square Error (RMSE,~\cite{willmott2005advantages}) is a quadratic score, in the same units of the dependent variable, in which the average error in the model predictions is computed by averaging the squared individual errors.
Our regression model achieves an RMSE equal to 7.76, representing the average difference between the actual ESG score and the inferred one.
Lastly, we report the model performance (7.9 \%) using the Weighted Mean Absolute Percentage Error (wMAPE,~\cite{kim2016new, hastie2009elements}), a scale-independent score that measures the average of absolute percentage errors. 

To conclude the review of the regression model performance, we conduct a residual analysis to check the linear assumptions required to properly shape the problem as a linear model.
The assumption of normal distribution of the residuals ($E_i \sim N(0, \sigma^2)$) is confirmed by the Anderson-Darling test~\cite{nelson1998anderson} with a $p$-value equal to 6.6 \% as well as through the QQ plot of residuals versus Normal distribution showing points lie on a line.
Concerning homoscedasticity, a condition in which the residual variance is constant across all the model predictions, there are no visible patterns in the scatter plot of residuals versus predicted ESG scores.
The same condition is confirmed by the scatter plot of the predicted ESG scores versus the actual scores. 
However, a slight overestimation trend might be spotted for ESG scores below 50, showing a limit of our predictors for interpreting these low scores. 
A graphical panel with all the graphical residual analyses is shown in the Supplementary Material document (see Section SM17).

\end{document}